\documentclass{article}

\usepackage{microtype}
\usepackage{graphicx}
\usepackage{subfigure}
\usepackage{booktabs} 
\usepackage{makecell}
\usepackage[ruled,vlined,linesnumbered]{algorithm2e}

\usepackage{hyperref}

\usepackage[accepted]{icml2025}

\usepackage{amsmath}
\usepackage{amssymb}
\usepackage{mathtools}
\usepackage{amsthm}
\usepackage{amsmath, amssymb}

\usepackage[capitalize,noabbrev]{cleveref}

\theoremstyle{plain}
\newtheorem{theorem}{Theorem}[section]

\newtheorem{lemma}[theorem]{Lemma}

\theoremstyle{definition}

\theoremstyle{remark}

\usepackage[textsize=tiny]{todonotes}

\usepackage{pifont}
\usepackage[linesnumbered,ruled,vlined]{algorithm2e}
\usepackage[table,xcdraw]{xcolor}
\usepackage{multirow}
\usepackage{adjustbox}
\usepackage{subcaption}
\usepackage{array}
\usepackage{diagbox}
\newcommand{\thickhline}{\noalign{\hrule height 1pt}}

\icmltitlerunning{Submission and Formatting Instructions for ICML 2025}

\begin{document}

\twocolumn[
\icmltitle{LION: A Clifford Neural Paradigm for Multimodal-Attributed Graph Learning}


\begin{icmlauthorlist}
\icmlauthor{Xunkai Li}{}
\icmlauthor{Zhengyu Wu}{}
\icmlauthor{Zekai Chen}{}
\icmlauthor{Henan Sun}{}
\icmlauthor{Daohan Su}{} 
\icmlauthor{Guang Zeng}{}
\icmlauthor{Hongchao Qin}{}
\icmlauthor{Rong-Hua Li}{}
\icmlauthor{Guoren Wang}{}
\end{icmlauthorlist}


\icmlcorrespondingauthor{Rong-Hua Li}{lironghuabit@126.com}

\icmlkeywords{Machine Learning, ICML}

\vskip 0.3in
]



\printAffiliationsAndNotice{}  

\begin{abstract}
    Recently, the rapid advancement of multimodal domains has driven a data-centric paradigm shift in graph ML, transitioning from text-attributed to multimodal-attributed graphs.
    This advancement significantly enhances data representation and expands the scope of graph downstream tasks, such as modality-oriented tasks, thereby improving the practical utility of graph ML.
    Despite its promise, limitations exist in the current neural paradigms:
    (1) Neglect Context in Modality Alignment: 
    Most existing methods adopt topology-constrained or modality-specific operators as tokenizers.
    These aligners inevitably neglect graph context and inhibit modality interaction, resulting in suboptimal alignment.
    (2) Lack of Adaptation in Modality Fusion: 
    Most existing methods are simple adaptations for 2-modality graphs and fail to adequately exploit aligned tokens equipped with topology priors during fusion, leading to poor generalizability and performance degradation.

    To address the above issues, we propose LION (c\underline{LI}ff\underline{O}rd \underline{N}eural paradigm) based on the Clifford algebra and decoupled graph neural paradigm (i.e., propagation-then-aggregation) to implement alignment-then-fusion in multimodal-attributed graphs. 
    Specifically, we first construct a modality-aware geometric manifold grounded in Clifford algebra.
    This geometric-induced high-order graph propagation efficiently achieves modality interaction, facilitating modality alignment.
    Then, based on the geometric grade properties of aligned tokens, we propose adaptive holographic aggregation. 
    This module integrates the energy and scale of geometric grades with learnable parameters to improve modality fusion.
    Extensive experiments on 9 datasets demonstrate that LION significantly outperforms SOTA baselines across 3 graph and 3 modality downstream tasks.
\end{abstract}

\section{Introduction}
\label{sec: Introduction}
    With the proliferation of multimodal data (e.g., images and texts) and vision-language models, multimodal-attributed graphs (MAGs) have recently emerged as a research frontier in the graph ML community. 
    In MAG, nodes denote real-world entities described by multimodal data, which substantially expands the semantic dimensionality of node features and elevates the upper bound of representation learning capability~\cite{yan2025magb,zhu2025mm_graph,liu2025graph_mllm,wang2025mg_llm}.
    This means that MAG not only offers immense potential for enhancing traditional graph-based tasks (e.g., node classification and link prediction) but also broadens the task spectrum to encompass modality-oriented retrieval and generation~\cite{ning2025graph4mm,fang2025graphgpt_o,jin2024instructg2i}, thereby enhancing practical utility.
    Consequently, developing effective MAG neural paradigms (i.e., MAGNNs) has become an urgent necessity.
    Despite the remarkable progress made by recent studies, some inherent limitations still exist. 
    Our in-depth analysis is as follows.

    \textbf{{Limitation: \!Neglect \!MAG-Context in \!Modality\! Alignment.}}
    We argue that most existing methods inherit unnecessary entity isolation, at the node and modality levels, from conventional graph-agnostic modality alignment strategies.
    Specifically, (1) some methods employ topology-constrained aligners~\cite{fan2025mlaga,fang2025graphgpt_o,ning2025graph4mm,jin2024instructg2i}. 
    While they aim to mitigate the receptive field limitations of the target nodes (i.e., single-entity) in MAG by incorporating broader node contexts, their scope remains restricted to 1-hop neighbors.
    Consequently, they still fail to capture the long-range dependencies, which have been proven to be highly beneficial in graphs~\cite{zhang2021ndls,li2024atp,liu2025sigma}.
    (2) While other methods explicitly leverage graph topology to facilitate global alignment, they remain restricted to modality-specific scenarios and rely on the same topology to apply identical alignment across different modalities~\cite{hu2025mig_gt,hu2025ntsformer,guo2025DMGC,he2025unigraph2}. 
    This coarse-grained alignment overlooks beneficial intra- and inter-modal interactions, thereby negatively impacting subsequent modality fusion.

    \textbf{Key Insight}.
    It is imperative to leverage modality-aware, high-quality topology to dissolve entity semantic boundaries in MAGs, thereby fostering beneficial, profound modality interactions and understanding for better alignment.

\begin{figure}[t]
  \centering
\setlength{\abovecaptionskip}{0.0cm}
\setlength{\belowcaptionskip}{-0.45cm}
  \includegraphics[width=\linewidth]{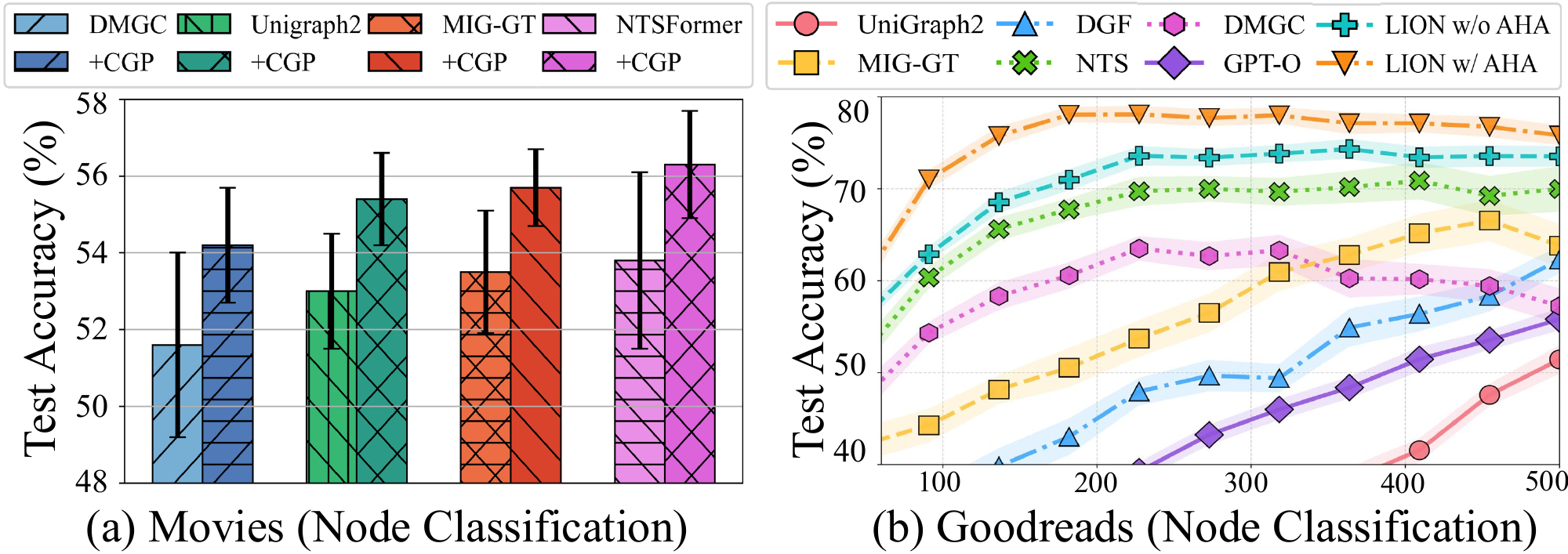}
  \vspace{-0.35cm}
  \caption{Empirical evaluations of CGP (a) and AHA (b), where the x-axis for AHA (b) is training time in seconds.}
  \label{fig: empirical}
\end{figure}

    \textbf{Solution and Evaluation}.
    To this end, we propose Clifford Geometric Propagation (CGP) to achieve efficient modality interaction for better alignment.
    Specifically, we first construct a MAG-specific geometric manifold where structural relationships are modeled as spatial rotations, and distinct modalities are represented as orthogonal geometric base vectors, rather than modality-specific scale vectors.
    Then, we introduce a geometric potential based on the graph topology to characterize semantic curvature as a principle for modality interaction.
    This formulation enables geometric-induced adaptive graph propagation (i.e., efficient modality interaction) and facilitates alignment.
    To evaluate CGP, we illustrate the performance gains in Fig.~\ref{fig: empirical}(a) by employing CGP as a plug-and-play module to replace existing modality aligners.
    Notably, despite some deeply coupled model architectures preventing extensive investigations, these 4 representative cases are indicative of the superiority of CGP.

    \textbf{{Limitation: Lack of MAG-Adaption in Modality Fusion.}}
    Given that enriched context, MAG significantly increases both the quantity and information density of aligned tokens, thereby presenting unique challenges.
    Specifically, (1) some methods directly employ Q-Former~\cite{li2023blip} to handle input scenarios involving topology-driven tokens (i.e., 1-hop neighbor tokens) by adaptive queries~\cite{fan2025mlaga,fang2025graphgpt_o,ning2025graph4mm,jin2024instructg2i}.
    Despite their simplicity and intuitiveness, these methods fail to explicitly leverage semantic-aware topology priors within the aligned tokens.
    (2) Although other methods attempt to enhance fusion by partitioning tokens into multiple topology- and modality-specific views and achieving fine-grained integration, such approaches offer only limited exploration of suboptimal aligned tokens and fail to deeply capture their intricate dependencies.~\cite{hu2025mig_gt,hu2025ntsformer,guo2025DMGC,he2025unigraph2,shen2025sr_gm,zheng2025dgf}.
    
    \textbf{Key Insight}.
    It is imperative to reveal the intricate dependency among aligned tokens and design a comprehensive fusion mechanism to unleash their representation potential.

\vspace{-0.1cm}
    \textbf{Solution and Evaluation}.
    After obtaining the propagated features in the geometric manifold (i.e., aligned tokens) via the CGP, we clarify that they naturally possess geometric grade properties, which encode the topology and modality insights.
    Based on this, we propose Adaptive Holographic Aggregation (AHA), which employs learnable parameters to explicitly model the energy and scale inherent in these geometric grades.
    This strategy functions as a dynamic filter, capturing the most relevant topology and modality insights for optimal fusion representation. 
    To substantiate our claims, we present comprehensive results in Fig.~\ref{fig: empirical}(b), where LION demonstrates superior convergence and performance compared to the SOTA baselines.
    Furthermore, the complete LION (i.e., LION w/ AHA) significantly outperforms the LION w/o AHA variant, thereby validating the advantages of modality fusion in the holographic perspective.

    \textbf{Contributions}.
    (1) \underline{\textit{New Perspective}}.
    We reveal the limitations of modality alignment and fusion in MAGs and introduce the mathematical framework grounded in Clifford algebra to address them for the first time.
    (2) \underline{\textit{Innovative Method}}.
    We propose LION (c\underline{LI}ff\underline{O}rd \underline{N}eural paradigm) based on this mathematical framework, which achieves alignment-then-fusion by propagation(CGP)-then-aggregation(AHA). 
    Specifically, CGP first constructs a MAG-specific geometric manifold to enable curvature-adaptive high-order graph propagation, achieving efficient modality interaction for better alignment. 
    Subsequently, AHA captures the energy and scale within the geometric grade properties of propagated features via learnable weights to facilitate modality fusion.
    (3) \underline{\textit{SOTA Performance}}.
    Evaluations from 6 domains demonstrate that LION achieves average improvements of 5.24\% and \!7.68\%\! over SOTA baselines in graph and modality tasks.

\section{Preliminaries}

\subsection{Notations and Problem Formulation}
    MAG is defined as $\mathcal{G}=(\mathcal{V}, \mathcal{E}, \{\mathbf{X}^{(m)}\}_{m\in\mathcal{M}})$ where $\mathcal{V}$ is the set of nodes, $\mathcal{E}$ is the set of edges, and $\mathcal{M}$ is the set of available modalities (e.g., texts and images). 
    For each node $v_i$ and modality $m$, modality-specific attribute vector ${x}_i^{(m)} \!\in \!\mathbb{R}^{d_m}$ constitutes complete $\mathbf{X} \!\in \!\mathbb{R}^{N \times d}$.
    The modality-shared topology is described by the same adjacency matrix $\mathbf{A}$, and $\mathbf{D}$ is the corresponding diagonal degree matrix.
    We further define the symmetric normalized graph Laplacian as $\tilde{\mathbf{L}} =\mathbf{I}-\tilde{\mathbf{D}}^{-1/2}\tilde{\mathbf{A}}\tilde{\mathbf{D}}^{-1/2}$ with self-loop ($\tilde{\mathbf{A}}=\mathbf{I}+\mathbf{A}$), which captures structural smoothness in graph signal processing.

    \textbf{Graph Tasks}.
    (1) Node Classification: predict the specific label class of unlabeled nodes; 
    (2) Link Prediction: predict if $(u,v)\in\mathcal{E}$ exists in the edge sets; 
    (3) Node Clustering: partition all nodes into clusters using off-the-shelf clustering algorithms applied to the learned representations.

    \textbf{Multimodal Tasks}. 
    (1) Modality Retrieval: given a query in a specific modality (e.g., text), retrieve the corresponding cross-modal representation;
    (2) Text Generation: generate text responses conditioned on the target node, instructions, and graph contexts; 
    (3) Image Generation: generate visual responses using diffusion models conditioned on the target node's textual descriptions and the graph contexts.

\subsection{MAG Neural Paradigm}
    \textbf{Conventional Methods.} 
    These approaches typically adapt previous multimodal architectures to MAGs by treating structural information as auxiliary tokens. 
    The core idea is linearizing neighborhoods into sequences~\cite{ning2025graph4mm,fang2025graphgpt_o} or projecting topology priors into soft tokens~\cite{fan2025mlaga,jin2024instructg2i}. 
    While leveraging pre-trained knowledge, these methods rely on concatenation or rigid alignment, lacking a unified integration where topology and modality are unified in a continuous space.

    \textbf{Graph-enhanced Methods.} 
    These approaches explicitly design specialized model architectures for MAGs—such as refined message-passing~\cite{hu2025mig_gt,hu2025ntsformer}, spectral graph filtering~\cite{shen2025sr_gm,guo2025DMGC}, and structure-conditioned modules~\cite{he2025unigraph2,zheng2025dgf}—to capture topology and modality insights. 
    However, these approaches often treat multimodal attributes and graph topology as separate semantic spaces, relying on well-designed but suboptimal alignment and fusion components.

\section{Methodology}
\label{sec: Methodology}
\subsection{Clifford Algebra and Geometric Manifold}
\label{sec: Clifford Algebra and Geometric Manifold}
    In this paper, we construct a mathematical framework specifically tailored for MAGs based on Clifford algebra, which extends the modality-specific graph operations to modality-aware high-dimensional geometric manifold spaces.
    Let $\mathcal{C}l_{n}$ denote the Clifford algebra defined over an $n$-dimensional space. 
    By equipping each node with a local tangent space isomorphic to $\mathcal{C}l_{n}$ and modeling edges as geodesic connections between them, we conceptualize the MAG as a discrete geometric manifold.
    The fundamental graph operation is the geometric product, which unifies spatial rotations (topology) with orthogonal geometric base vectors (modality).
    
    Specifically, for any two connected nodes $u, v$ with attribute vectors $x_u, x_v$, their topology-aware modality interaction (i.e., transform modality-aware geometric base vectors via spatial rotations in the geometric manifolds) is formalized as $x_u x_v \!=\! x_u \cdot x_v \!+\! x_u \wedge x_v$.
    This geometric product explicitly decomposes the multimodal interaction:
    (1) The symmetric inner product (${x}_u \!\cdot {x}_v$) induces a modality-specific projection, facilitating intra-modality interaction.
    (2) The antisymmetric outer product ($x_u \!\wedge\! x_v$) instantiates a bi-vector plane that encodes the topology curvature induced by cross-modality discrepancies, facilitating inter-modality interaction.
    This geometric manifold, spanned by Clifford algebra, provides a mathematical framework for topology-aware intra- and inter-modality interaction. 
    Based on this, it enables modality alignment through topology-driven, curvature-induced spatial rotations (i.e., graph propagation) and facilitates modality fusion by capturing the geometric grade properties of aligned tokens (i.e., message aggregation).

\subsection{Clifford Geometric Propagation}
\label{sec: Clifford Geometric Propagation}
\vspace{-0.04cm}
\textbf{Motivation.}
    In order to achieve efficient modality alignment via the CGP (i.e., modality-aware, curvature-adaptive high-order graph propagation within the geometric manifold), it is essential to capture topology insights from the curved manifold to facilitate multimodal interactions.
    Consequently, we introduce parallel transport, which aligns the distinct semantic spaces of multiple modalities within connected nodes by rotating modality-aware orthogonal geometric basis vectors along the topology curvature.
    This high-dimensional transformation preserves interaction channels across arbitrary modalities and is mathematically formulated via the spatial rotor $\mathcal{R}$ and geometric potential $\Phi$.
    Collectively, they modulate the principles of intra- and inter-modality interaction within a modality-aware geometric manifold.
    As a norm-preserving isometry, they ensure signal energy conservation, achieving efficient modality alignment for MAGs.

\textbf{Modality-oriented Clifford Initialization.}
    To begin with, we need to initialize $\mathbf{X}$ to support CGP.
    To achieve this, we lift the raw scalar attributes of $K$ modalities into an additional $2^K$-dimensional geometric manifold spanned by the Clifford algebra $\mathcal{C}l_K$. 
    Each node $u$ is equipped with a local tangent space where the initial multi-vector representation $\mathbf{H}_u^{(0)}$ is constructed by assigning each modality $k$ to an orthogonal Grade-1 basis vector $e_k\in\mathcal{C}l_K$. 
    This isomorphic lifting map is defined as:
    \begin{equation}
    \label{eq: modality-oriented Clifford initialization}
        \begin{aligned}
            \!\mathbf{H}_u^{(0)}\!\!\in\!\mathbb{R}^{d\times2^K}\!\!\coloneqq\! \psi \!\left( \sum_{k \in \mathcal{M}} {x}_{u}^{(k)} {e}_{k}\! \right)\!,\;\text{s.t. }\! \left\|\mathbf{H}_u^{(0)}\right\|_{\mathcal{C}l} \!=\! 1,
        \end{aligned}
    \end{equation}
    where $\psi(\cdot)$ is the normalization function.
    This constructed manifold is partitioned into $K+1$ geometric grades, where each grade ($d$) is a specialized channel for modality interaction.
    Specifically, the scalar subspace (Grade-0) captures the projections of identical modalities.
    The bi-vector subspace (Grade-2) instantiates interaction planes between modality pairs.
    For $K\ge 3$, the higher-grade subspaces facilitate multi-way interactions.
    In this paper, we assume $K=2$ by default to introduce the subsequent computational details.

\textbf{Topology-oriented \!Geometric \!Potential.}
    After that, we aim to capture intricate semantic curvature among $\mathbf{H}\coloneqq\mathbf{H}^{(0)}$ to facilitate modality interactions in Grade-0 and Grade-2.
    To this end, we propose the modality-adaptive geometric potential to modulate the semantic curvature between connected nodes, thereby equipping modality-adaptive properties within $\mathbf{A}$.
    Specifically, each edge $(u, v)$ is modeled as a geodesic parameterized by modality distributions. 
    Based on this, we derive the modality-adaptive geometric potential via the geometric product introduced in Sec.~\ref{sec: Clifford Algebra and Geometric Manifold}, which encapsulates both intra- (the scalar term) and inter-modality interaction (the bi-vector term) as follows:
    \begin{equation}
    \begin{aligned}
    \label{eq: topology-oriented Clifford initialization}
    \!\Phi_{uv} = \exp \left( - \frac{\|\langle \mathbf{H}_u \mathbf{H}_v \rangle_2\|_{\mathcal{C}l}^2}{\|\langle \mathbf{H}_u \mathbf{H}_v \rangle_0\|_{\mathcal{C}l} + \epsilon} \right),\;\forall (u,v)\in\mathcal{E},
    \end{aligned}
    \end{equation}
    where the scalar term $\| \langle \cdot \rangle_{0} \|_{\mathcal{C}l}$ measures the same semantic projection coverage, supplemented by an infinitesimal term $\epsilon$ for numerical stability.
    $\| \langle \cdot \rangle_{2} \|^2_{\mathcal{C}l}$ measures the magnitude of the bi-vector term, which geometrically interprets semantic curvature as the area of the interaction plane spanned by cross-modality disparities.
    $\exp(\cdot)$ acts as a decay kernel, mapping the intra- and inter-modality interaction ratio into a normalized weight interval.
    This strategy transcends the traditional homophily assumption and structural smoothing by leveraging modality-aware orthogonal basis vectors within a high-dimensional geometric manifold, thereby fully preserving the interaction channels across arbitrary modalities and reflecting the principles of multimodal interactions.
    
\textbf{Training-free Geometric Propagation.}
    Based on the above formulas, we extend conventional graph propagation into the high-order geometric manifold space, generalizing message passing by potential-gated parallel transport.
    This improved propagation unifies intra- and inter-modality interaction into a single coherent operation without any neural parameters, enabling efficient one-time computation on CPUs. 
    Finally, we formally define this alignment (graph propagation) process as follows:
    \begin{equation}
    \label{eq: Training-free Geometric Propagation}
    \begin{aligned}
    &\mathbf{H}_{u}^{(l)}\!\!\in\!\mathbb{R}^{d\times2^K}\!\!\!\coloneqq\! \mathbf{H}_u^{(l-1)} \!+\!\!\!\!\!\sum_{{(u, v) \in \mathcal{E}}} \!\!\!\!\tilde{\Phi}_{uv}\! \left(\! \mathcal{R}_{uv} \mathbf{H}_v^{(l-1)} \mathcal{R}_{uv}^{-1} \!\right),
    \\&\;\;\;\;\;\mathcal{R}_{uv} \!=\! \exp\!\left(\!-\frac{\!\langle \mathbf{H}_u \mathbf{H}_v\rangle_{2} }{2\| \langle \mathbf{H}_u \mathbf{H}_v\rangle_{2} \|_{\mathcal{C}l}^{2}} \right)\!,\;\forall (u,v)\in\mathcal{E},
    \end{aligned}
    \end{equation}
    where $\mathcal{R}$ and $\Phi$ both emerge from the curvature bi-vector within the geometric manifold space.
    Collectively, $\mathcal{R}_{uv}$ and normalized $\tilde{\Phi}_{uv}$ characterize the principles of intra- and inter-modal interactions among connected nodes, thereby aligning the semantic orientation of $v$ with the local tangent space of $u$ via the rotation of modality-aware orthogonal basis vectors.
    It ensures that high-order graph propagation is strictly guided by the topology curvature of the geometric manifold to achieve efficient modality alignment.

\textbf{Theoretical Analysis.}
    We first establish the geometric stability bound of Clifford high-dimensional manifold space (Theorem~\ref{theorem1} and Appendix~\ref{appendix: Proof of Geometric Stability}).
    Based on this, we demonstrate that the parallel-transport-optimized message passing inherently achieves modality alignment via Clifford Dirichlet energy minimization (Theorem~\ref{theorem2} and Appendix~\ref{appendix: Proof of Spectral Evidence}).
    \begin{theorem}
    \label{theorem1}
    \textbf{(Stability Bound of Clifford Manifold).}
    Let $f$ be the mapping function that obtains $\!\mathbf{H}$ and $\mathcal{A_G}\!\coloneqq\!\{\Phi,\mathcal{R}\}$ in the Clifford geometric manifold. 
    Assuming the constituent operations satisfy Lipschitz continuity, for any paired input $(\mathbf{X},\mathbf{A})$ subject to bounded perturbations, the divergence within the constructed manifold is strictly bounded by:
\begin{equation}
\label{eq: theorem1}
    \begin{aligned}
        &\;\;\;\;\| \mathbf{H} - \mathbf{H}' \|_{\mathcal{C}l} + \| \mathcal{A}_{\mathcal{G}} - \mathcal{A}'_{\mathcal{G}} \|_{\mathcal{C}l} \le\\ 
        &K_{\text{map}} \cdot \left( \|\mathbf{X} - \mathbf{X}'\|_{\mathcal{C}l} + \gamma \|\mathbf{A} - \mathbf{A}'\|_{\mathcal{C}l} \right),
    \end{aligned}
\end{equation}
    where $K_{\text{map}}$ and $\gamma$ are the Lipschitz constant determined strictly by the mathematical properties of Clifford operators.
\end{theorem}

\begin{theorem}
\label{theorem2}
\!\!\textbf{(Spectral Evidence for Modality \!Alignment).}
    The parallel-transport-optimized message passing essentially constitutes a gradient descent optimization that minimizes the potential-gated Clifford Dirichlet energy, denoted as $\mathbb{E}_{\text{Dir}}(\mathbf{H})$. 
    Unlike traditional scalar smoothing that strictly relies on homophily, $\mathbb{E}_{\text{Dir}}(\mathbf{H})$ is governed jointly by the geometric potential and the curvature-induced spatial rotor:
    \begin{equation}
    \label{eq: theorem2}
    \mathbb{E}_{\text{Dir}}(\mathbf{H}) \coloneqq \frac{1}{2} \sum_{(u, v) \in \mathcal{E}} \underbrace{\Phi_{uv} \left\| \mathbf{H}_u - \mathcal{R}_{uv} \mathbf{H}_v \mathcal{R}_{uv}^{-1} \right\|_{\mathcal{C}l}^2}_{\text{Curvature-driven Geometric Misalignment}}.
    \end{equation} 
    Theoretically, the alignment error decays exponentially at a rate of $(1 - \lambda_{\min}(\tilde{\mathcal{L}}_{\mathcal{G}}))$, where $\tilde{\mathcal{L}}_{\mathcal{G}}$ is the potential-induced geometric Laplacian. 
    It provides convergence guarantees while preserving all modality interaction channels.
    \end{theorem}

\subsection{Adaptive Holographic Aggregation}
\label{sec: Adaptive Holographic Aggregation}
\textbf{Motivation.} 
    To unleash the representation potential of geometric propagated features (i.e., aligned tokens), it is imperative to reveal their intricate dependencies.
    Unfortunately, prevalent message aggregation (e.g., simple concat or mean pooling) in scalar space indiscriminately compresses geometric channels and ignores receptive fields, leading to poor generalizability and performance degradation.  
    To address this issue, we clarify that they naturally exhibit geometric grade properties within the Clifford manifold, where specific grades encode distinct topology and modality insights (e.g., scalars for intra-modality, bi-vectors for inter-modality interactions).  
    Based on this, we propose AHA, a holographic-inspired dynamic filter that explicitly models the energy and scale of geometric grades for optimal fusion representation.

\textbf{Energy-aware Grade Filtering.}
    In summary, we aim to disentangle and selectively prioritize informative modality interaction channels from the propagated features. 
    To achieve this, we define the grade energy $\mathbb{E}_{u}^{(l)}\in\mathbb{R}^{2^K}$ containing the squared Clifford norms of each grade component to quantify the information density.
    Then, we explicitly map the Grade-0 and Grade-2 into 2 intra- and 2 inter-modality interaction channels:
    \begin{equation}
    \label{eq: energy_energy}
    \begin{aligned}
        \mathbb{E}_{u}^{(l)} \!=\! \left[ \left\|\langle \mathbf{H}_u^{(l)} \rangle_{0}\right\|_{\mathcal{C}l}^2,\left\|\langle \mathbf{H}_u^{(l)} \rangle_{0}\right\|_{\mathcal{C}l}^2,., \left\|\langle \mathbf{H}_u^{(l)} \rangle_{2}\right\|_{\mathcal{C}l}^2 \right]^\top\!\!\!\!.
    \end{aligned}
    \end{equation}
    After this initialization, we introduce a learnable energy gate parameterized by $\mathbf{W}_{\mathcal{G}}$ to dynamically modulate these channels, yielding interaction-specific scores $\boldsymbol{\alpha}_{u}^{(l)}$ for filtering:
    \begin{equation}
    \label{eq: energy_gating}
    \begin{aligned}
        \!\tilde{\mathbf{H}}_u^{(l)} \!=\! \mathbf{H}_u^{(l)} \!\odot\! \boldsymbol{\alpha}_{u}^{(l)},\;
        \boldsymbol{\alpha}_{u}^{(l)} \!= \!\sigma\! \left(\! \mathbf{W}_{\mathcal{G}} \!\cdot\! \text{Norm}(\mathbb{E}_{u}^{(l)}) \!+\! \mathbf{b}_{\mathcal{G}}\! \right)\!,
    \end{aligned}
\end{equation}
    where $\odot$ is element-wise product.
    This strategy functions as a task-adaptive spectral filter for the propagated features within the Clifford manifold space.
    Specifically, although it initially prioritizes high-energy components, it simultaneously incorporates low-energy components as required by grounding the adaptive gating mechanism in $\mathbb{E}_{u}^{(l)}$.

\textbf{Scale-aware Resonance Fusion.}
    After obtaining weighted modality interaction channels, their validity varies across topology scales corresponding to the propagation depth $L$ within each channel.
    To reconcile these receptive fields and facilitate fusion, we first generate $\mathbf{H}_u^{\text{ctx}}$ by learnable $\mathbf{W}_{\tau}$ to capture the consensus profile across all scales:
\begin{equation}
\label{eq: scale_weight}
\begin{aligned}
    \!\mathbf{H}_u^{\text{ctx}}\in\mathbb{R}^{d\times2^K}\! \!\coloneqq\! \text{Norm} \left(   \sum_{l=0}^{L} \left[\mathbf{W}_{\tau}^{(l)}\tilde{\mathbf{H}}_u^{(l)}+\! \mathbf{b}_{\tau}^{(l)}\right] \!  \right).
\end{aligned}
\end{equation}
    Subsequently, we employ a resonance attention mechanism to compute the scale validity score $\beta_{u,l}$, where a learnable attention score $\mathbf{a}_S^\top$ projects the interaction between the current representation and consensus profile into a scalar space.
    Based on this, AHA aggregates the grade-filtered and scale-weighted tokens, and the $d_f$-dimension fusion representation $\mathbf{Z}_u$ is obtained by projecting it back into Euclidean space via a Clifford linear layer parameterized by $\mathbf{W}_{\text{out}}$:
\begin{equation}
\label{eq: scale_aggregation}
\begin{aligned}
    &\;\;\;\mathbf{Z}_u\in\!\mathbb{R}^{d_f}\coloneqq \mathbf{W}_{\text{out}} \left( \sum_{l=0}^{L} \beta_{u,l} \cdot \tilde{\mathbf{H}}_u^{(l)} \right) + \mathbf{b}_{\text{out}},\\
    &\beta_{u,l} = \frac{\exp(\mathbf{a}_S^\top \cdot \tanh(\mathbf{W}_S [\mathbf{H}_u^{\text{ctx}} \| \tilde{\mathbf{H}}_u^{(l)}]))}{\sum_{k=0}^{L} \exp(\mathbf{a}_S^\top \cdot \tanh(\mathbf{W}_S [\mathbf{H}_u^{\text{ctx}} \| \tilde{\mathbf{H}}_u^{(k)}]))}.
\end{aligned}
\end{equation}

\textbf{Theoretical Analysis.}
    We establish the theoretical guarantee of AHA by formulating the fusion as a holographic reconstruction problem. 
    Unlike scalar aggregation, which assumes uniform information density, AHA reconstructs the optimal fusion representation by minimizing the risk associated with grade-wise noise and scale variance. 
    In Theorem~\ref{theorem3} and Appendix~\ref{appendix: Proof of Holographic Reconstruction}, we rigorously bound this error to demonstrate the convergence properties of our method.
    \begin{theorem}
        \label{theorem3}
        \textbf{(Holographic Reconstruction Error Bound).}
         Let $\mathbf{Z}^*$ be the optimal fusion representation for the specific downstream task and $\mathbf{N}_u$ denote the manifold noise.
         The reconstruction error of the aggregated representation $\mathbf{Z}_u$ is strictly bounded by the residual energy of the filtered grades and the divergence from the geometric consensus profile:
         \vspace{-0.05cm}
         \begin{equation}
         \label{eq: theorem3}
         \begin{aligned}
            \left\| \mathbf{Z}_u - \mathbf{Z}^* \right\| \le &\sum_{l=0}^{L} \beta_{u,l} \underbrace{\left\| (\mathbf{1} - \boldsymbol{\alpha}_u^{(l)}) \odot \mathbf{N}_u^{(l)} \right\|}_{\text{Noise Suppression}} \\
            &+ \sum_{l=0}^{L} \beta_{u,l}\underbrace{\omega \left\| \tilde{\mathbf{H}}_u^{(l)} - \mathbf{H}_u^{\text{ctx}} \right\|}_{\text{Scale Consensus}},
        \end{aligned}
        \end{equation}
        where $\omega$ is the Lipschitz constant associated with manifold curvature. 
        The first term indicates that energy-aware grade filtering ($\boldsymbol{\alpha}$) minimizes the noise upper bound by suppressing irrelevant geometric grades. 
        The second term demonstrates that scale-aware resonance fusion ($\beta$) reduces structural deviation, ensuring consensus profile convergence.
    \end{theorem}
    Detailed algorithm descriptions, key insights, and complexity analysis of LION are provided in Appendix~\ref{appendix: Algorithm and Complexity Analysis}.

\section{Experiments}
\label{sec: experiments}
    In this section, we first introduce the experimental setup, and additional details for reproducibility are provided in the Appendix~\ref{appendix: Dataset Description}-\ref{appendix: Experiment Environment}. 
    Then, we provide a comprehensive evaluation to address the following questions:
    \textbf{Q1 (Effectiveness)}. 
    How does LION perform as a new neural paradigm for MAG?
    \textbf{Q2 (Ablation)}. 
    If LION is effective, what contributes to its performance? 
    \textbf{Q3 (Interpretability)}. 
    In depth, how do these components exert their influence?
    \textbf{Q4 (Robustness)}. 
    How robust is LION when dealing with sparse scenarios?
    \textbf{Q5 (Scalability)}. 
    What are the time overhead, space overhead, and convergence efficiency of LION?
\subsection{Experimental Setup}

\textbf{Datasets}.
    In our experiments, we evaluate LION across 9 publicly available MAG datasets from 6 domains, achieving a comprehensive validation. 
    Specifically, these datasets include social network (RedditS)~\cite{desai1_RedditS}, movie network (Movies)~\cite{ni2019_Grocery_Cloth_Ele_Movies_Sports}, 4 recommendation networks (Grocery, Sports, Ele-fashion, Cloth)~\cite{ni2019_Grocery_Cloth_Ele_Movies_Sports,hou2024_Cloth_Ele_Sports}, art network (SemArt)~\cite{garcia2018_SemArt}, image network (Flickr30k)~\cite{plummer2015_Flickr30k}, and book network (Goodreads)~\cite{wan2018_Goodreads_NC,wan2019_Goodreads_NC}. 
    Due to space constraints, the statistics and description details are summarized in Appendix~\ref{appendix: Dataset Description}.

\textbf{Baselines}.
    To achieve a comprehensive comparison, we utilize 
    (i) Single-modality GNN: GCN, GAT, GCNII, GATv2;
    (ii) Simple MAGNN: MMGCN, MGAT;
    (iii) Conventional MAGNN: MLaGA, GraphGPT-O, Graph4MM, InstructG2I;
    (iv) Graph-enhanced MAGNN: DMGC, DGF, MIG-GT, NTSFormer, UniGraph2. 
    Details are shown in Appendix~\ref{appendix: Baselines Details}.

\textbf{Evaluation Protocols}.
    Fundamentally, the research motivation for MAG is to enhance data quality for traditional graph tasks and expand the graph downstream task spectrum to improve practical utility. 
    Therefore, evaluation involves
    (i) Prevalent graph tasks: node classification, link prediction, node clustering;
    (ii) Novel graph-oriented modality tasks: modality retrieval, Graph-to-Text (G2Text), Graph-to-Image (G2Image). 
    Given the complexity of the evaluation pipelines, hyperparameter settings, and quantitative metrics, please refer to Appendix~\ref{appendix: Evaluation Protocols} for comprehensive details.

\subsection{Performance Comparison}
\label{sec: Performance Comparison}

\begin{table*}[htbp]
\setlength{\abovecaptionskip}{0.2cm}
\setlength{\belowcaptionskip}{-0.2cm}
\centering
\caption{Performance comparison on graph downstream tasks.
The best/second result is \textbf{bold}/\underline{underline}.}
\label{tab: graph downstream performance comparison}
\footnotesize 
\renewcommand{\arraystretch}{1.1}
\resizebox{\linewidth}{!}{
\setlength{\tabcolsep}{1.5mm}{
\begin{tabular}{l!{\vrule width 0.1pt}cc!{\vrule width 0.1pt}cc!{\vrule width 0.1pt}cc!{\vrule width 0.1pt}cc!{\vrule width 0.1pt}cc!{\vrule width 0.1pt}cc}
\hline\thickhline
\rowcolor{gray!80}
\multicolumn{1}{c!{\vrule width 0.1pt}}{\textbf{\textcolor{white}{Tasks}}} & \multicolumn{4}{c!{\vrule width 0.1pt}}{\textbf{\textcolor{white}{Node Classification}}} & \multicolumn{4}{c!{\vrule width 0.1pt}}{\textbf{\textcolor{white}{Link Prediction}}} & \multicolumn{4}{c}{\textbf{\textcolor{white}{Node Clustering}}} \\
\hline
\rowcolor{gray!20}
 & \multicolumn{2}{c!{\vrule width 0.1pt}}{\textbf{Movies}} & \multicolumn{2}{c!{\vrule width 0.1pt}}{\textbf{Goodreads}} & \multicolumn{2}{c!{\vrule width 0.1pt}}{\textbf{Sports}} & \multicolumn{2}{c!{\vrule width 0.1pt}}{\textbf{Cloth}} & \multicolumn{2}{c!{\vrule width 0.1pt}}{\textbf{RedditS}} & \multicolumn{2}{c}{\textbf{Grocery}} \\
\cline{2-13}
\rowcolor{gray!20}
\multirow{-2}{*}{\diagbox[width=8.5em,height=2.4em]{\textbf{Methods}}{\textbf{Datasets}}} 
 & Acc & F1-Score & Acc & F1-Score & MRR & Hits@3 & MRR & Hits@3 & NMI & ARI & NMI & ARI \\
\hline
GCN & 39.12$_{ \pm 1.33}$ & 36.17$_{ \pm 1.41}$ & 56.03$_{ \pm 1.52}$ & 50.14$_{ \pm 1.21}$ & 46.23$_{ \pm 0.28}$ & 58.09$_{ \pm 0.73}$ & 40.52$_{ \pm 0.37}$ & 48.15$_{ \pm 0.44}$ & 62.18$_{ \pm 0.62}$ & 64.07$_{ \pm 0.51}$ & 38.13$_{ \pm 0.43}$ & 29.54$_{ \pm 0.38}$ \\
\rowcolor{gray!10} MMGCN & 44.87$_{ \pm 1.42}$ & 39.92$_{ \pm 1.37}$ & 58.54$_{ \pm 0.87}$ & 52.36$_{ \pm 1.53}$ & 48.44$_{ \pm 0.34}$ & 60.52$_{ \pm 0.58}$ & 42.83$_{ \pm 0.41}$ & 50.94$_{ \pm 0.56}$ & 72.10$_{ \pm 0.55}$ & 70.08$_{ \pm 0.62}$ & 45.84$_{ \pm 0.49}$ & 32.26$_{ \pm 0.41}$ \\
GAT & 40.25$_{ \pm 1.48}$ & 36.43$_{ \pm 1.52}$ & 57.12$_{ \pm 1.08}$ & 51.08$_{ \pm 1.24}$ & 47.19$_{ \pm 0.43}$ & 59.14$_{ \pm 0.77}$ & 41.26$_{ \pm 0.49}$ & 49.34$_{ \pm 0.58}$ & 65.45$_{ \pm 0.68}$ & 66.28$_{ \pm 0.63}$ & 39.42$_{ \pm 0.45}$ & 28.75$_{ \pm 0.47}$ \\
\rowcolor{gray!10} MGAT & 43.14$_{ \pm 1.37}$ & 38.09$_{ \pm 1.44}$ & 59.88$_{ \pm 0.91}$ & 53.75$_{ \pm 1.38}$ & 49.82$_{ \pm 0.36}$ & 62.03$_{ \pm 0.54}$ & 43.91$_{ \pm 0.45}$ & 52.54$_{ \pm 0.51}$ & 73.33$_{ \pm 0.53}$ & 69.14$_{ \pm 0.59}$ & 43.08$_{ \pm 0.43}$ & 31.15$_{ \pm 0.46}$ \\
\hline
MLaGA & 48.37$_{ \pm 1.02}$ & 42.08$_{ \pm 1.64}$ & 66.42$_{ \pm 1.25}$ & 60.29$_{ \pm 1.13}$ & 54.33$_{ \pm 0.62}$ & 68.41$_{ \pm 0.88}$ & 49.12$_{ \pm 0.73}$ & 58.37$_{ \pm 0.80}$ & 82.58$_{ \pm 1.15}$ & 82.14$_{ \pm 1.12}$ & 51.92$_{ \pm 0.83}$ & 37.58$_{ \pm 0.89}$ \\
\rowcolor{gray!10} GraphGPT-O & 49.12$_{ \pm 1.90}$ & 42.94$_{ \pm 1.13}$ & 64.25$_{ \pm 0.92}$ & 58.76$_{ \pm 1.30}$ & 55.82$_{ \pm 0.68}$ & 69.17$_{ \pm 0.75}$ & 51.43$_{ \pm 0.66}$ & 60.18$_{ \pm 0.73}$ & 79.33$_{ \pm 1.02}$ & 80.41$_{ \pm 0.96}$ & 50.64$_{ \pm 0.69}$ & 38.83$_{ \pm 0.62}$ \\
Graph4MM & 49.76$_{ \pm 1.85}$ & 43.22$_{ \pm 1.95}$ & 67.18$_{ \pm 0.78}$ & 61.35$_{ \pm 0.83}$ & 53.94$_{ \pm 0.71}$ & 67.58$_{ \pm 0.83}$ & 50.84$_{ \pm 0.68}$ & 59.43$_{ \pm 0.63}$ & 84.14$_{ \pm 0.92}$ & 83.25$_{ \pm 0.90}$ & 51.20$_{ \pm 0.78}$ & 38.92$_{ \pm 0.75}$ \\
\rowcolor{gray!10} InstructG2I & 50.57$_{ \pm 1.74}$ & 44.08$_{ \pm 1.38}$ & 65.84$_{ \pm 0.90}$ & 59.93$_{ \pm 1.02}$ & 56.75$_{ \pm 0.83}$ & 70.12$_{ \pm 0.93}$ & 52.28$_{ \pm 0.80}$ & 61.54$_{ \pm 0.86}$ & 82.83$_{ \pm 1.09}$ & 81.58$_{ \pm 1.05}$ & 49.87$_{ \pm 0.92}$ & 35.28$_{ \pm 0.82}$ \\
\hline
DMGC & 51.73$_{ \pm 1.95}$ & 46.84$_{ \pm 1.59}$ & 63.42$_{ \pm 1.27}$ & 61.18$_{ \pm 1.63}$ & 56.47$_{ \pm 0.54}$ & 69.83$_{ \pm 0.49}$ & 51.87$_{ \pm 0.61}$ & 60.84$_{ \pm 0.57}$ & \underline{89.62}$_{ \pm 0.73}$ & \underline{89.14}$_{ \pm 0.80}$ & \underline{56.13}$_{ \pm 0.66}$ & 41.82$_{ \pm 0.60}$ \\
\rowcolor{gray!10} DGF & 52.94$_{ \pm 1.64}$ & 45.13$_{ \pm 1.70}$ & 66.83$_{ \pm 1.61}$ & 62.54$_{ \pm 1.12}$ & 55.93$_{ \pm 0.51}$ & 69.17$_{ \pm 0.54}$ & 51.24$_{ \pm 0.63}$ & 60.13$_{ \pm 0.59}$ & 88.87$_{ \pm 0.70}$ & 88.42$_{ \pm 0.76}$ & {55.88}$_{ \pm 0.63}$ & \underline{42.25}$_{ \pm 0.68}$ \\
MIG-GT & 53.17$_{ \pm 2.02}$ & \underline{47.60}$_{ \pm 1.89}$ & {68.24}$_{ \pm 0.90}$ & {64.87}$_{ \pm 1.24}$ & 59.13$_{ \pm 0.47}$ & 71.94$_{ \pm 0.44}$ & {55.12}$_{ \pm 0.56}$ & \underline{63.24}$_{ \pm 0.53}$ & 86.25$_{ \pm 0.82}$ & 84.89$_{ \pm 0.88}$ & 51.62$_{ \pm 0.70}$ & 38.18$_{ \pm 0.66}$ \\
\rowcolor{gray!10} NTSFormer & \underline{53.89}$_{ \pm 2.16}$ & {46.94}$_{ \pm 1.93}$ & 71.19$_{ \pm 0.93}$ & \underline{65.80}$_{ \pm 0.80}$ & \underline{60.52}$_{ \pm 0.44}$ & \underline{72.28}$_{ \pm 0.41}$ & 54.83$_{ \pm 0.53}$ & 62.94$_{ \pm 0.50}$ & 85.81$_{ \pm 0.78}$ & 84.14$_{ \pm 0.83}$ & 52.33$_{ \pm 0.68}$ & 37.81$_{ \pm 0.63}$ \\
UniGraph2 & 52.92$_{ \pm 1.79}$ & 46.28$_{ \pm 1.66}$ & \underline{72.82}$_{ \pm 0.82}$ & {65.32}$_{ \pm 0.96}$ & 59.84$_{ \pm 0.49}$ & 71.26$_{ \pm 0.47}$ & \underline{55.31}$_{ \pm 0.54}$ & 63.18$_{ \pm 0.52}$ & 86.56$_{ \pm 0.68}$ & 85.24$_{ \pm 0.73}$ & 54.42$_{ \pm 0.64}$ & 40.86$_{ \pm 0.62}$ \\
\rowcolor{gray!10} {LION (Ours)} & \textbf{58.61}$_{ \pm 1.28}$ & \textbf{51.73}$_{ \pm 1.75}$ & \textbf{78.54}$_{ \pm 0.70}$ & \textbf{68.91}$_{ \pm 1.13}$ & \textbf{62.31}$_{ \pm 0.55}$ & \textbf{73.87}$_{ \pm 0.39}$ & \textbf{58.47}$_{ \pm 0.85}$ & \textbf{66.58}$_{ \pm 0.48}$ & \textbf{90.53}$_{ \pm 0.89}$ & \textbf{90.17}$_{ \pm 1.08}$ & \textbf{58.54}$_{ \pm 0.72}$ & \textbf{46.12}$_{ \pm 0.89}$ \\
\hline\thickhline
\end{tabular}
}}
\end{table*}

\begin{table}[htbp]
\setlength{\abovecaptionskip}{0.2cm}
\setlength{\belowcaptionskip}{-0.2cm}
\centering
\caption{Performance comparison on modality-level tasks.}
\label{tab: multimodal downstream performance comparison}
\footnotesize 
\renewcommand{\arraystretch}{1.1}
\resizebox{\linewidth}{!}{
\setlength{\tabcolsep}{1.2mm}{
\begin{tabular}{l!{\vrule width 0.1pt}cc!{\vrule width 0.1pt}cc!{\vrule width 0.1pt}cc}
\hline\thickhline
\rowcolor{gray!80}
\multicolumn{1}{c!{\vrule width 0.1pt}}{\textbf{\textcolor{white}{Tasks}}} & \multicolumn{2}{c!{\vrule width 0.1pt}}{\textbf{\textcolor{white}{Modal Retrieval}}} & \multicolumn{2}{c!{\vrule width 0.1pt}}{\textbf{\textcolor{white}{G2Text}}} & \multicolumn{2}{c}{\textbf{\textcolor{white}{G2Image}}} \\
\hline

\rowcolor{gray!20}
 & \multicolumn{2}{c!{\vrule width 0.1pt}}{\textbf{Ele-fashion}} & \multicolumn{2}{c!{\vrule width 0.1pt}}{\textbf{Flickr30k}} & \multicolumn{2}{c}{\textbf{SemArt}} \\
\cline{2-7}

\rowcolor{gray!20}
\multirow{-2}{*}{\diagbox[width=8em,height=2.4em]{\textbf{Methods}}{\textbf{Datasets}}} 
 & \;MRR\; & Hits@3
 & BLEU-4 & \;CIDEr\;
 & \;CLIP-S\; & DINOv2-S \\
\hline

GCNII & 77.24 & 70.15 & 5.42 & 39.54 & 50.52 & 35.81 \\
\rowcolor{gray!10} MMGCN & 81.85 & 74.58 & 6.15 & 44.82 & 54.88 & 39.87 \\
GATv2 & 78.53 & 71.32 & 5.56 & 40.24 & 51.25 & 36.54 \\
\rowcolor{gray!10} MGAT & 82.41 & 75.26 & 6.28 & 45.56 & 55.43 & 40.52 \\
\hline
MLaGA & 87.65 & 79.85 & 9.54 & 71.58 & 68.52 & 53.15 \\
\rowcolor{gray!10}  GraphGPT-O & 88.45 & 80.50 & 9.89 & 72.58 & \underline{70.84} & 54.29 \\
Graph4MM & 86.25 & 78.42 & \underline{10.42} & \underline{74.82} & 67.21 & 52.56 \\
\rowcolor{gray!10}  InstructG2I & 89.12 & 81.24 & 9.68 & 71.98 & 69.92 & \underline{55.43} \\
\hline
DMGC & 91.20 & 83.12 & 7.83 & 61.82 & 61.54 & 47.26 \\
\rowcolor{gray!10}  DGF & 91.55 & 83.47 & 7.92 & 62.20 & 61.98 & 47.68 \\
MIG-GT & 92.54 & 84.51 & 8.15 & 63.57 & 63.82 & 48.94 \\
\rowcolor{gray!10}  NTSFormer & 92.88 & 84.93 & 8.24 & 64.12 & 63.26 & 48.56 \\
Unigraph2 & \underline{93.13} & \underline{85.45} & 8.56 & 65.28 & 64.56 & 49.83 \\
\rowcolor{gray!10} {LION (Ours)} & \textbf{94.67} & \textbf{86.92} & \textbf{11.54} & \textbf{78.92} & \textbf{74.21} & \textbf{58.30} \\
\hline\thickhline
\end{tabular}
}}
\vspace{-0.6cm}
\end{table}

\textbf{Graph Tasks}.
    To answer \textbf{Q1}, we first report performance comparison on graph tasks.
    As shown in Table~\ref{tab: graph downstream performance comparison}, LION consistently achieves superior performance across all datasets, tasks, and metrics.
    Specifically, in node classification, LION outperforms the leading NTSFormer and Unigraph2 by a significant average margin of 5.84\%.
    This success highlights the capacity of LION to capture topology and modality insights through the Clifford geometric manifold.
    
    Regarding the baselines, graph-enhanced methods like NTSFormer generally outperform conventional methods such as MLaGA, yet they remain inferior to LION primarily due to their reliance on suboptimal Euclidean spaces.
    Furthermore, most approaches fail to maintain consistent dominance across all three tasks and six metrics. 
    For instance, while clustering-specific DMGC and DGF exhibit competitiveness performance, they fail to generalize effectively to the node classification task.
    Meanwhile, there are currently no specialized models designed for the link-level task.

\textbf{Modality Tasks}.
    Table~\ref{tab: multimodal downstream performance comparison} further validates the superiority of LION in modality tasks, encompassing prevalent retrieval and generation, where it achieves SOTA performance across all evaluation metrics.
    A notable performance gain is observed on the SemArt dataset for the G2Image task, where LION outperforms the leading baselines, GraphGPT-O and InstructG2I, by 4.76\% in CLIP-S and 5.18\% in DINOv2-S. 
    These results demonstrate that our alignment-then-fusion MAG neural paradigm effectively preserves the semantic integrity of multimodal attributes.
    This strategy facilitates more coherent and accurate content generation compared to existing methods that treat modality and topology as separate spaces.
    Notably, modality-agnostic prevalent GNNs and early MMGNNs fail to achieve competitive performance due to their outdated model architecture designs.

\begin{table}[htbp]
\setlength{\abovecaptionskip}{0.2cm}
\setlength{\belowcaptionskip}{-0.2cm}
\centering
\caption{Ablation Study.}
\label{tab: ablation_study}
\renewcommand{\arraystretch}{1.1}
\resizebox{\linewidth}{!}{
\setlength{\tabcolsep}{0.8mm}{
\begin{tabular}{l!{\vrule width 0.1pt}cc!{\vrule width 0.1pt}cc!{\vrule width 0.1pt}cc!{\vrule width 0.1pt}cc}
\hline\thickhline
\rowcolor{gray!80}
\multicolumn{1}{c!{\vrule width 0.1pt}}{\textbf{\textcolor{white}{Tasks}}} & \multicolumn{4}{c!{\vrule width 0.1pt}}{\textbf{\textcolor{white}{Node Classification}}} & \multicolumn{4}{c}{\textbf{\textcolor{white}{Link Prediction}}} \\
\hline
\rowcolor{gray!20}
 & \multicolumn{2}{c!{\vrule width 0.1pt}}{\textbf{RedditS}} & \multicolumn{2}{c!{\vrule width 0.1pt}}{\textbf{Grocery}} & \multicolumn{2}{c!{\vrule width 0.1pt}}{\textbf{Flickr30k}} & \multicolumn{2}{c}{\textbf{SemArt}} \\
\cline{2-9}
\rowcolor{gray!20}
\multirow{-2}{*}{\diagbox[width=8em,height=2.4em]{\textbf{Methods}}{\textbf{Datasets}}} 
 & Acc & F1 & Acc & F1 & MRR & Hits@3 & MRR & Hits@3 \\
\hline
LION (Ours) & \textbf{94.8$_{ \pm 0.4}$} & \textbf{90.6$_{ \pm 0.5}$} & \textbf{88.3$_{ \pm 0.7}$} & \textbf{78.1$_{ \pm 0.9}$} & \textbf{68.4$_{ \pm 0.5}$} & \textbf{76.5$_{ \pm 0.4}$} & \textbf{85.2$_{ \pm 0.3}$} & \textbf{89.6$_{ \pm 0.2}$} \\
\rowcolor{gray!10} w/o Rotor & 93.9$_{ \pm 0.3}$ & 89.4$_{ \pm 0.4}$ & 87.4$_{ \pm 0.6}$ & 77.6$_{ \pm 0.8}$ & 67.8$_{ \pm 0.5}$ & 76.0$_{ \pm 0.4}$ & 84.7$_{ \pm 0.3}$ & 89.1$_{ \pm 0.2}$ \\
w/o Potential & 93.4$_{ \pm 0.4}$ & 88.7$_{ \pm 0.5}$ & 86.6$_{ \pm 0.7}$ & 76.8$_{ \pm 0.8}$ & 67.2$_{ \pm 0.4}$ & 75.7$_{ \pm 0.5}$ & 84.3$_{ \pm 0.3}$ & 88.8$_{ \pm 0.2}$ \\
\rowcolor{gray!10} w/o Energy & 92.8$_{ \pm 0.5}$ & 88.3$_{ \pm 0.6}$ & 86.2$_{ \pm 0.8}$ & 75.9$_{ \pm 1.0}$ & 66.5$_{ \pm 0.6}$ & 74.5$_{ \pm 0.5}$ & 83.5$_{ \pm 0.4}$ & 87.5$_{ \pm 0.3}$ \\
w/o Consen. & 92.6$_{ \pm 0.4}$ & 88.5$_{ \pm 0.6}$ & 85.9$_{ \pm 0.7}$ & 76.6$_{ \pm 0.9}$ & 67.0$_{ \pm 0.5}$ & 74.8$_{ \pm 0.5}$ & 82.9$_{ \pm 0.4}$ & 87.9$_{ \pm 0.3}$ \\
\rowcolor{gray!10} w/o Scale & 91.7$_{ \pm 0.6}$ & 86.8$_{ \pm 0.7}$ & 84.5$_{ \pm 0.9}$ & 75.5$_{ \pm 1.1}$ & 66.8$_{ \pm 0.7}$ & 73.5$_{ \pm 0.6}$ & 82.5$_{ \pm 0.5}$ & 86.7$_{ \pm 0.4}$ \\

\hline\thickhline
\rowcolor{gray!80}
\multicolumn{1}{c!{\vrule width 0.1pt}}{\textbf{\textcolor{white}{Tasks}}} & \multicolumn{4}{c!{\vrule width 0.1pt}}{\textbf{\textcolor{white}{Node Clustering}}} & \multicolumn{4}{c}{\textbf{\textcolor{white}{Multimodal Retrieval}}} \\
\hline
\rowcolor{gray!20}
 & \multicolumn{2}{c!{\vrule width 0.1pt}}{\textbf{Movies}} & \multicolumn{2}{c!{\vrule width 0.1pt}}{\textbf{Ele-Fashion}} & \multicolumn{2}{c!{\vrule width 0.1pt}}{\textbf{Sports}} & \multicolumn{2}{c}{\textbf{Cloth}} \\
\cline{2-9}
\rowcolor{gray!20}
\multirow{-2}{*}{\diagbox[width=8em,height=2.4em]{\textbf{Methods}}{\textbf{Datasets}}} 
 & NMI & ARI & NMI & ARI & MRR & Hits@3 & MRR & Hits@3 \\
\hline
LION (Ours) & \textbf{24.6$_{ \pm 0.4}$} & \textbf{10.6$_{ \pm 0.4}$} & \textbf{56.8$_{ \pm 0.6}$} & \textbf{48.9$_{ \pm 0.8}$} & \textbf{95.2$_{ \pm 0.2}$} & \textbf{88.7$_{ \pm 0.1}$} & \textbf{92.5$_{ \pm 0.1}$} & \textbf{89.1$_{ \pm 0.2}$} \\
\rowcolor{gray!10} w/o Rotor & 23.7$_{ \pm 0.3}$ & 10.0$_{ \pm 0.3}$ & 56.0$_{ \pm 0.5}$ & 47.9$_{ \pm 0.7}$ & 94.3$_{ \pm 0.2}$ & 88.1$_{ \pm 0.1}$ & 91.8$_{ \pm 0.1}$ & 88.6$_{ \pm 0.2}$ \\
w/o Potential & 23.5$_{ \pm 0.4}$ & 9.4$_{ \pm 0.3}$ & 55.9$_{ \pm 0.5}$ & 48.1$_{ \pm 0.7}$ & 93.9$_{ \pm 0.2}$ & 87.9$_{ \pm 0.1}$ & 90.5$_{ \pm 0.1}$ & 87.8$_{ \pm 0.2}$ \\
\rowcolor{gray!10} w/o Energy & 22.8$_{ \pm 0.5}$ & 8.9$_{ \pm 0.4}$ & 54.6$_{ \pm 0.7}$ & 46.9$_{ \pm 0.9}$ & 93.8$_{ \pm 0.3}$ & 87.0$_{ \pm 0.2}$ & 90.9$_{ \pm 0.2}$ & 87.1$_{ \pm 0.3}$ \\
w/o Consen. & 23.2$_{ \pm 0.4}$ & 8.7$_{ \pm 0.3}$ & 54.8$_{ \pm 0.6}$ & 46.5$_{ \pm 0.8}$ & 93.5$_{ \pm 0.3}$ & 86.4$_{ \pm 0.2}$ & 90.4$_{ \pm 0.2}$ & 87.3$_{ \pm 0.3}$ \\
\rowcolor{gray!10} w/o Scale & 22.9$_{ \pm 0.5}$ & 8.2$_{ \pm 0.2}$ & 54.2$_{ \pm 0.8}$ & 45.9$_{ \pm 1.0}$ & 93.0$_{ \pm 0.4}$ & 86.3$_{ \pm 0.3}$ & 90.1$_{ \pm 0.3}$ & 86.8$_{ \pm 0.4}$ \\

\hline\thickhline
\end{tabular}
}}
\end{table}

\subsection{Ablation Study}
\label{sec: Ablation Study}
    To answer \textbf{Q2}, we conduct an ablation study to investigate the contribution of each module within LION.  
    As detailed in Sec.~\ref{sec: Methodology}, LION consists of two core modules, namely CGP and AHA. 
    To evaluate their impact, we define five module variants. 
    For CGP in Eq.~(\ref{eq: Training-free Geometric Propagation}), (1) w/o Rotor and (2) w/o Potential remove the spatial rotor $\mathcal{R}$ and geometric potential $\Phi$ . 
    For AHA in Eq.~(\ref{eq: energy_gating}), Eq.~(\ref{eq: scale_weight}), and Eq.~(\ref{eq: scale_aggregation}), (3) w/o Energy, (4) w/o Consen., and (5) w/o Scale omit the energy-aware gating, consensus profile, and scale-aware resonance fusion.

    Based on this, Table~\ref{tab: ablation_study} confirms the necessity of module design. 
    Regarding the CGP, removing the spatial rotor and geometric potential leads to a significant performance drop. 
    This verifies that modeling topology as geometric rotations and differentiating interaction principles via semantic curvature is crucial for effective alignment. 
    Regarding the AHA, the w/o Energy variant exhibits poor performance. 
    This demonstrates that simply aggregating all geometric grades introduces redundancy and confirms that filtering based on information density is critical. 
    Furthermore, the performance degradation in w/o Scale and w/o Consen. confirms that adaptive fusion guided by the consensus profile allows the model to optimally reconcile receptive fields, which is particularly beneficial for multiple interaction channels.

\begin{figure*}[t]
\setlength{\abovecaptionskip}{0.2cm}
\setlength{\belowcaptionskip}{-0.2cm}
  \includegraphics[width=\textwidth]{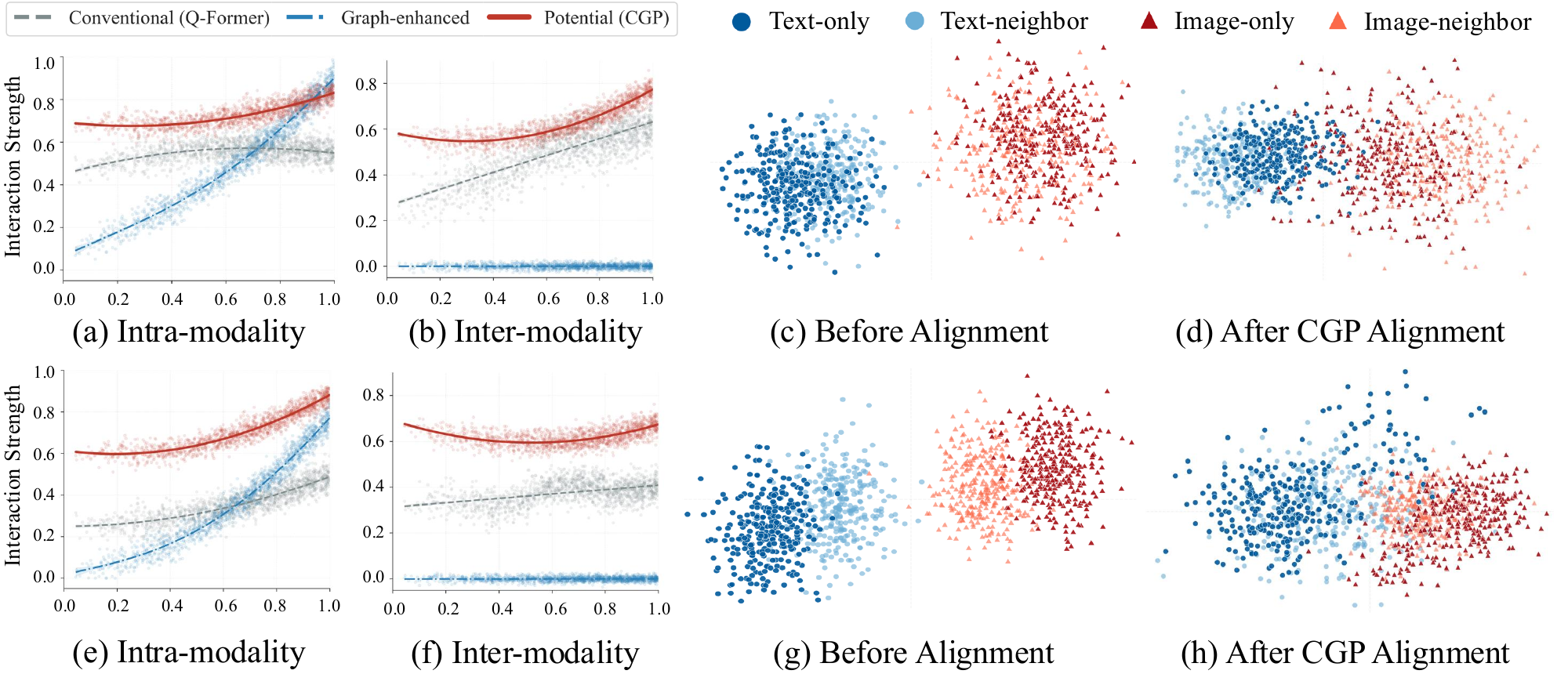}
  \caption{(a)-(d) The G2Text visualization of CGP modality interaction and alignment (i.e., geometric-induced adaptive graph propagation) on Flickr30k.
  (e)-(h) The G2Image visualization of CGP modality interaction and alignment on SemArt. 
  }
  \label{fig: cgp_alignment}
\end{figure*}

\begin{figure}[t]
  \centering
\setlength{\abovecaptionskip}{0.2cm}
\setlength{\belowcaptionskip}{-0.2cm}
  \includegraphics[width=\linewidth]{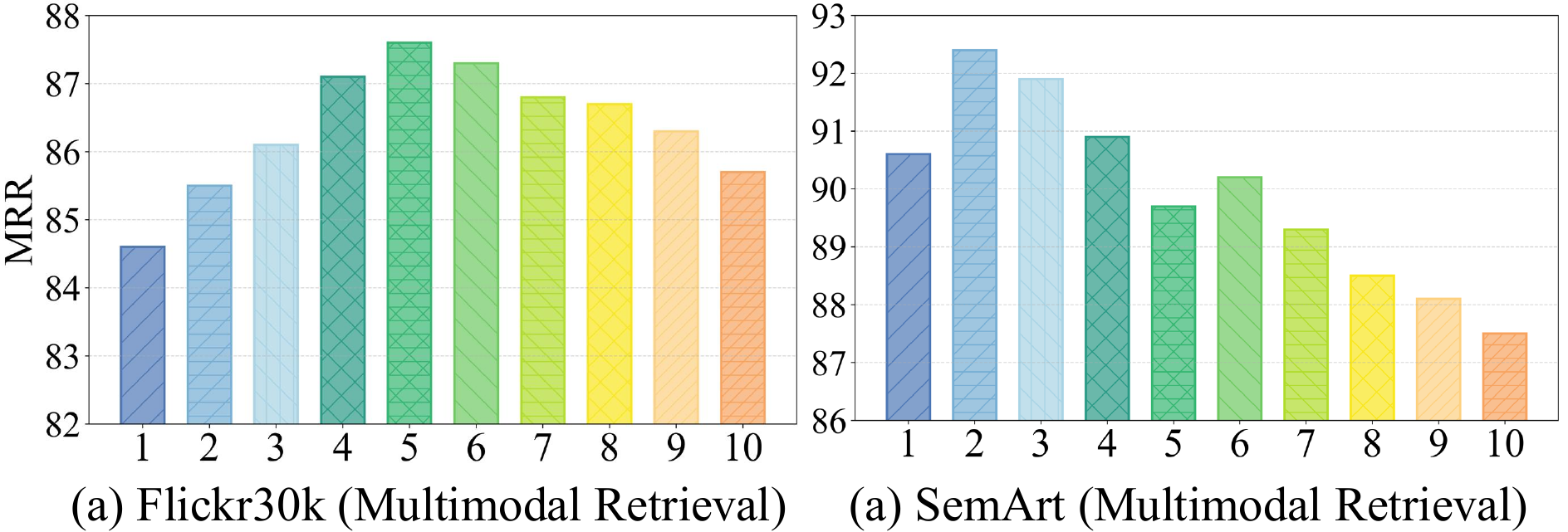}
  \caption{Hyperparameter analysis of CGP depth.}
  \label{fig: cgp_depth}
\end{figure}

\begin{figure}[t]
  \centering
\setlength{\abovecaptionskip}{0.2cm}
\setlength{\belowcaptionskip}{-0.2cm}
  \includegraphics[width=\linewidth]{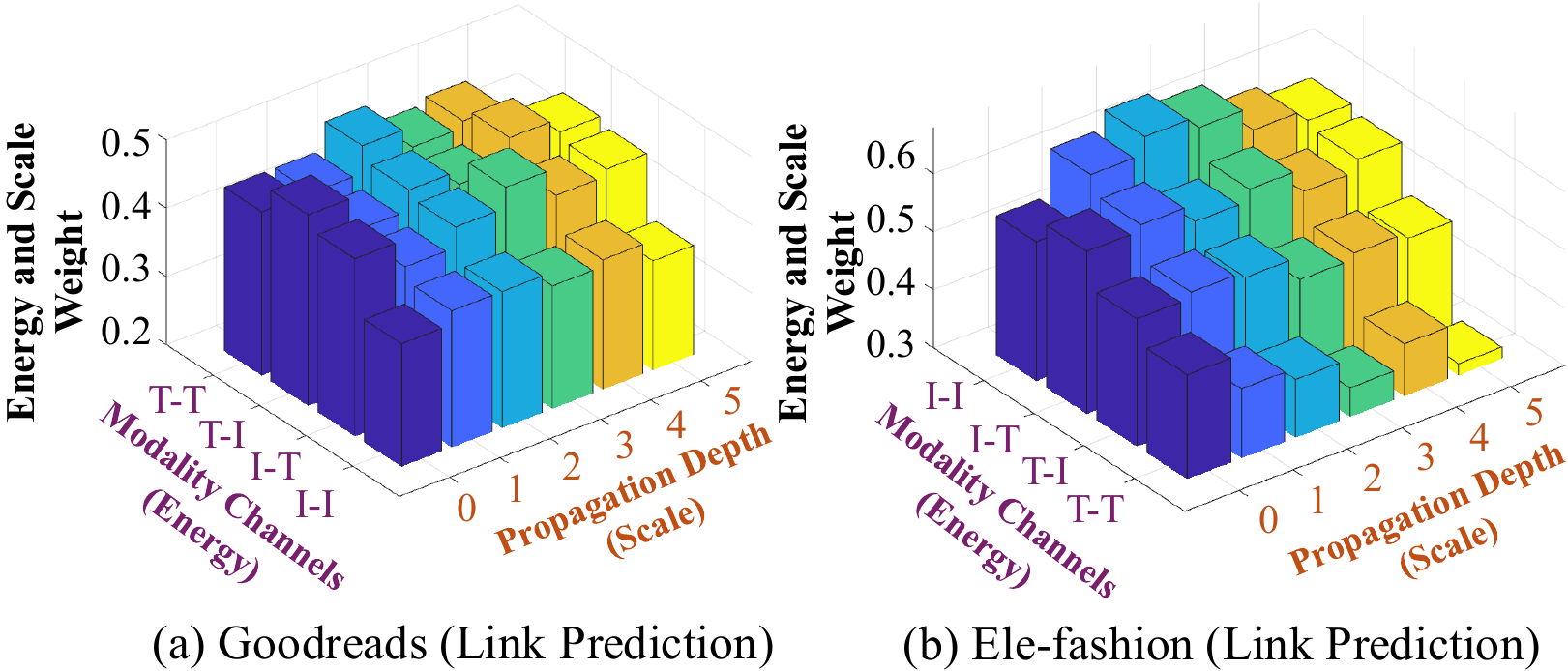}
  \caption{AHA visualization.
  T and I are text and image.}
  \label{fig: aha_fusion}
  \vspace{-0.2cm}
\end{figure}

\subsection{In-depth Analysis}
\label{sec: In-depth Analysis}

\textbf{CGP Module}.
    To answer \textbf{Q3}, we provide a visual analysis to intuitively interpret the underlying mechanisms of the CGP and AHA modules.
    First of all, we plot the intra- and inter-modality interaction strength (y-axis) against connected node feature similarity (x-axis). 
    As evident in Fig.~\ref{fig: cgp_alignment} (a)-(b) for Flickr30k and Fig.~\ref{fig: cgp_alignment} (e)-(f) for SemArt, the modality interaction baselines exhibit a sharp decay in interaction strength as node feature similarity decreases, which reveals a heavy reliance on the homophily assumption.
    Notably, the modality interactions in graph-enhanced methods occur exclusively within modality-specific views, resulting in an inter-modality interaction strength of zero.
    
    In contrast, CGP breaks the node homophily assumption by modality-aware geometric manifolds and topology-adaptive high-order graph propagation. 
    Specifically, by harmonizing modality interaction principles (Sec.~\ref{sec: Clifford Geometric Propagation}), CGP effectively facilitates both intra- and inter-modality interactions.
    Based on this, we visualize the token latent space before and after alignment to demonstrate that this curvature-aware multimodality interaction reshapes the representation and ultimately benefits downstream tasks. 
    
    Prior to alignment, the latent spaces of modalities appear disjointed. 
    After CGP, a distinct trend emerges where entity representations transcend modality boundaries to capture high-level semantic dependencies. 
    This strategy effectively gathers semantically related modalities while distancing others.
    Meanwhile, G2Text and G2Image tasks exhibit enhanced text and image compactness, reflecting an adaptive alignment that conforms to the specific downstream requirements.
    This result intuitively demonstrates that our spatial rotor and geometric potential successfully guide the parallel transport of modality attribute vectors for efficient alignment.
    Furthermore, Fig.~\ref{fig: cgp_depth} illustrates the impact of varying CGP propagation depths on model performance. 
    Experimental results suggest that a smaller propagation depth is preferable for dense graphs, while sparse graphs benefit from increased depth to capture a sufficient receptive field.

\begin{figure*}[t]
\setlength{\abovecaptionskip}{0.2cm}
\setlength{\belowcaptionskip}{-0.2cm}
  \includegraphics[width=\textwidth]{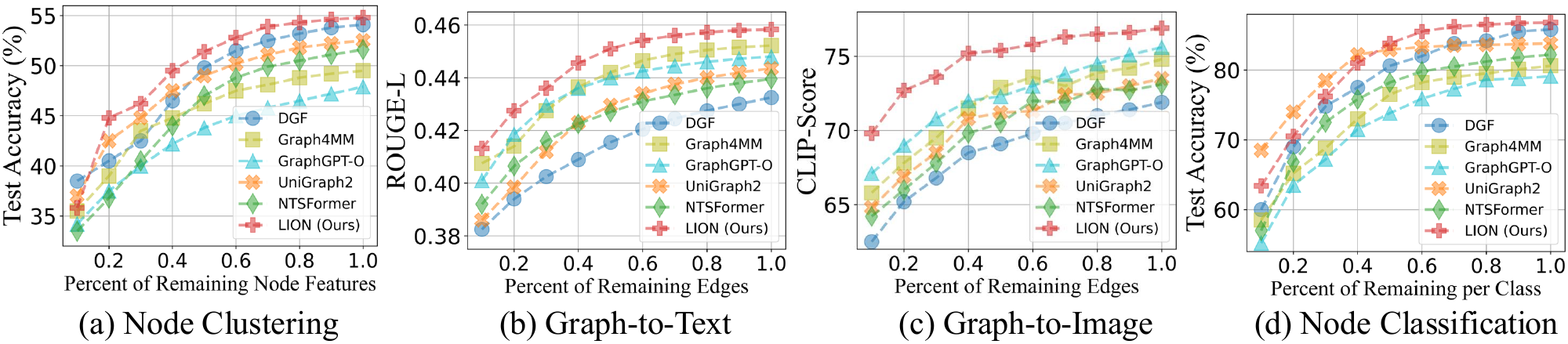}
  \caption{The model performance of LION on the sparsity Grocery.
  }
  \label{fig: robust_sparsity}
\end{figure*}

\begin{table}[htbp]
\setlength{\abovecaptionskip}{0.2cm}
\setlength{\belowcaptionskip}{-0.2cm}
\centering
\caption{Epoch-Batch efficiency on Goodreads (Classifi.).}
\label{tab: efficiency}
\footnotesize 
\renewcommand{\arraystretch}{1.1}
\resizebox{\linewidth}{!}{
\setlength{\tabcolsep}{1.5mm}{
\begin{tabular}{l!{\vrule width 0.1pt}c!{\vrule width 0.1pt}c!{\vrule width 0.1pt}c!{\vrule width 0.1pt}c!{\vrule width 0.1pt}c}
\hline\thickhline
\rowcolor{gray!80}
\textbf{\textcolor{white}{Method}} & \textbf{\textcolor{white}{Pre-process (s)}} & \textbf{\textcolor{white}{E-Train. (s)}} & \textbf{\textcolor{white}{E-Infer. (s)}} & \textbf{\textcolor{white}{B-GPU Mem.}} & \textbf{\textcolor{white}{Param.}} \\
\hline
MLaGA       & - & 112.4$_{\pm 1.05}$ & 45.2$_{\pm 0.93}$ & 22.8G & 125M \\
\rowcolor{gray!10} GraphGPT-O  & - & 125.6$_{\pm 1.70}$ & 48.9$_{\pm 1.45}$ & 24.5G & 140M \\
Graph4MM    & 45.2$_{\pm 1.03}$ & 28.5$_{\pm 0.85}$ & 10.4$_{\pm 0.42}$ & 17.2G & 96M \\
\rowcolor{gray!10} InstructG2I & - & 168.2$_{\pm 2.50}$ & 75.6$_{\pm 1.10}$ & 32.1G & 180M \\
\hline
DMGC        & - & 9.2$_{\pm 0.15}$ & 4.1$_{\pm 0.08}$ & 3.8G & 2.4M \\
\rowcolor{gray!10} DGF         & - & 10.8$_{\pm 0.20}$ & 3.6$_{\pm 0.10}$ & 4.1G & 2.6M \\
MIG-GT      & 63.7$_{\pm 1.16}$ & 5.4$_{\pm 0.12}$ & 2.5$_{\pm 0.05}$ & 2.2G & 1.8M \\
\rowcolor{gray!10} NTSFormer   & 75.8$_{\pm 1.25}$ & 4.1$_{\pm 0.10}$ & 1.9$_{\pm 0.04}$ & 2.5G & 2.1M \\
{LION (Ours)} & 92.4$_{\pm 1.49}$ & {3.2}$_{\pm 0.09}$ & {1.2}$_{\pm 0.04}$ & {2.9G} & {1.5M} \\
\hline\thickhline
\end{tabular}
}}
\end{table}

\begin{figure}[t]
  \centering
\setlength{\abovecaptionskip}{0.2cm}
\setlength{\belowcaptionskip}{-0.2cm}
  \includegraphics[width=\linewidth]{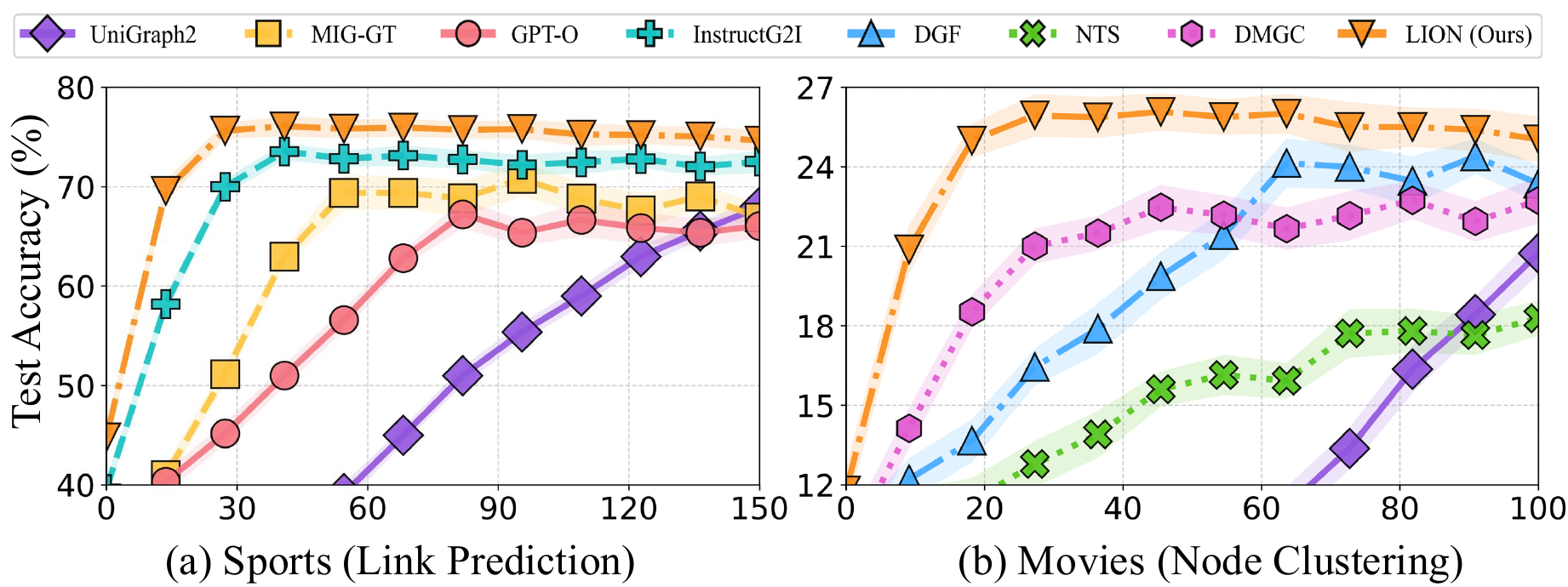}
  \caption{The convergence efficiency curves.}
  \label{fig: convergence}
  \vspace{-0.2cm}
\end{figure}

\textbf{AHA Module}.
    First of all, we clarify the modality fusion mechanism implemented from a holographic perspective involving both energy and scale (Sec.~\ref{sec: Adaptive Holographic Aggregation}).
    In Fig.~\ref{fig: aha_fusion}, modality channels are primarily filtered via energy-initialized adaptive gating to select the informative and beneficial interaction subspaces. 
    Then, propagation depth determines the manner in which multi-scale structural contexts are integrated within these interaction channels.
    Based on this, we reveal that LION dynamically assigns varying importance to channels (energy) and depth (scale) according to the specific scenarios. 
    For instance, a distinct modality-aware preference emerges where the model assigns significantly higher weights to text-related subspaces on Goodreads while shifting its focus toward image-related subspaces on Ele-Fashion. 
    Meanwhile, their optimal receptive fields also vary across these datasets to ensure optimal modality fusion.

\subsection{Robustness Analysis}
\label{sec: Robustness Analysis}
    To answer \textbf{Q4}, we investigate the robustness of LION in sparse scenarios.
    This stems from the fact that MAGs often suffer from data incompleteness.
    As shown in Fig.~\ref{fig: robust_sparsity}, LION demonstrates remarkable robustness across all sparsity scenarios. 
    Meanwhile, LION consistently outperforms other methods in all downstream tasks. 
    This superiority is attributed to the Clifford geometric manifold, which can mitigate missing data through high-order graph propagation. 
    This indicates that our alignment-then-fusion mechanism effectively exploits the complementary nature of topology and modality, allowing the model to learn high-quality representations even when raw information is severely corrupted.

\subsection{Running Efficiency}
\label{sec: Running Efficiency}
    To answer \textbf{Q5}, we provide the computational overhead and convergence curves of LION.
    As summarized in Table~\ref{tab: efficiency}, LION demonstrates superior efficiency compared to other baselines. 
    This advantage is primarily attributed to our decoupled graph neural paradigm, which separates intensive geometric computations from the training loop. 
    Specifically, the construction of the spatial rotor and geometric potential within the CGP is relegated to a one-time pre-processing.
    Consequently, the training phase focuses exclusively on the AHA and avoids the need for recursive neighborhood expansion.
    Fig.~\ref{fig: convergence} illustrates that LION exhibits a rapid convergence rate, consistently reaching peak performance in significantly fewer epochs than other baselines. 
    This result aligns with our theoretical analysis in Theorem~\ref{theorem2}, where we proved that the potential-gated energy minimization inherently acts as a geometric regularization. 
    This mechanism effectively constrains the optimization path by penalizing structurally inconsistent updates, thereby smoothing the loss landscape. 
    By reducing the complexity of the neural search space, LION facilitates accelerated adaptation.

\section{Conclusion}
    With the rapid advancement of multimodal domains, MAG has emerged as a critical frontier for enhancing graph representation and downstream utility. 
    In this work, we propose LION, which unifies topology with modality by constructing the geometric manifold grounded in Clifford algebra. 
    Our theoretical analysis confirms that LION constitutes a rigorous optimization process, thereby ensuring its efficacy.
    Extensive experiments across a broad spectrum of supervised and unsupervised tasks demonstrate that LION significantly outperforms SOTA baselines, confirming its superiority in both graph and
    modality tasks.
    Although the manifold dimension scales exponentially with the number of modalities, this overhead remains marginal in practice, as the modality count is typically limited.
    Future work includes the integration of LION into broader applications such as Graph QA. 
    Furthermore, enhancing the utility of Multimodal Large Models within LION to fully exploit cross-modal semantics under topology priors is also a compelling direction.

\newpage

\section*{Impact Statement}
This paper presents work whose goal is to advance the field of Machine
Learning.
There are many potential societal consequences of our work, none of which we feel must be specifically highlighted here.

\bibliography{example_paper}
\bibliographystyle{icml2025}

\newpage
\appendix
\onecolumn

\section{Proof of Geometric Stability in Theorem~\ref{theorem1}}
\label{appendix: Proof of Geometric Stability}
    In this section, we provide the formal mathematical proof for the stability of the Clifford manifold construction as asserted in Theorem~\ref{theorem1}.
    Complementing the theoretical analysis in the main text, we rigorously examine the geometric transformation process employed by LION.
    Our primary objective is to demonstrate that the mapping function $f$ satisfies the Lipschitz continuity condition.
    This function projects raw modality attributes and graph topology into Clifford multi-vectors and geometric operators.
    Formally, we prove that for any perturbation in the input space, the deviation in the resulting Clifford representation is bounded by a constant factor of that perturbation.
    Establishing this continuity is fundamental to the theoretical validity of our framework.
    It ensures the manifold is inherently robust to input noise and guarantees that minor feature fluctuations or structural anomalies do not lead to catastrophic divergence in the latent geometric embeddings.

\begin{proof}
    Let the input space be defined by a Clifford geometric manifold, into which the node modality attribute vectors $\mathbf{X} \in \mathbb{R}^{N \times d}$ and the adjacency matrix $\mathbf{A}$ are embedded. 
    We subsequently consider perturbed inputs $\mathbf{X}'$ and $\mathbf{A}'$ such that $\|\mathbf{X} - \mathbf{X}'\|_{\mathcal{C}l} \le \epsilon_{\mathbf{X}}$ and $\|\mathbf{A} - \mathbf{A}'\|_{\mathcal{C}l} \le \epsilon_{\mathbf{A}}$. 
    The mapping function $f$, as defined in Sec.~\ref{sec: Clifford Geometric Propagation}, comprises the following two primary components for the Clifford adaptation of modality attribute vectors $\mathbf{X}$ and the topology $\mathbf{A}$: 
    (1) Modality-oriented Clifford Initialization, where the lifting map $\psi$ embeds the raw attributes $\mathbf{X}$ into the multi-vector space $\mathbf{H}$; 
    (2) Topology-oriented Geometric Potential and Spatial Rotor, where the high-order graph propagation operator derives the Clifford-valued adjacency $\mathcal{A}_{\mathcal{G}}$ 
    Based on this, we define the total distance $\mathcal{D}_{\mathcal{M}}$ within the Clifford geometric manifold as the sum of the modality attribute divergence and the graph topology divergence:
\begin{equation}
\begin{aligned}
    \mathcal{D}_{\mathcal{M}} = \|\mathbf{H} - \mathbf{H}'\|_{\mathcal{C}l} + \|\mathcal{A}_{\mathcal{G}} - \mathcal{A}'_{\mathcal{G}}\|_{\mathcal{C}l}.
\label{eq:appendix_manifold_divergence}
\end{aligned}
\end{equation}

\begin{lemma}
\label{theorem_proof_lemma1}
    (Stability of Modality-oriented Clifford Initialization). 
    Recall the modality-oriented Clifford initialization function: $\mathbf{H}_u^{(0)} = \sum_{k=1}^{K} \psi(\mathbf{x}_u^{(k)}) \mathbf{e}_k$, where $\psi(\mathbf{x}_u^{(k)}) = \mathbf{x}_u^{(k)}/\|\mathbf{x}_u^{(k)}\|_2$. 
    Assuming modality attribute vectors are bounded away from zero, the normalization $\psi(\cdot)$ is Lipschitz continuous with constant $L_{\psi}$. 
    Since the direct sum with orthogonal basis vectors $\mathbf{e}_k$ is a linear isometry, we have:
    \begin{equation}
    \begin{aligned}
        \|\mathbf{H} - \mathbf{H}'\|_{\mathcal{C}l} \le L_{\psi} \|\mathbf{X} - \mathbf{X}'\|_{\mathcal{C}l} = \delta_H.
    \label{eq:lemma1_bound}
    \end{aligned}
    \end{equation}
    Crucially, given that each modality component is unit-normalized, the Clifford norm of the resulting multi-vector is bounded by $\|\mathbf{H}_u\|_{\mathcal{C}l} = \sqrt{\sum_{k=1}^K 1^2} = \sqrt{K}$.
\end{lemma}

\begin{lemma}
\label{theorem_proof_lemma2}
    (Stability of Topology-oriented Geometric Potential and Spatial Rotor). 
    The Clifford geometric adjacency matrix $\mathcal{A}_{\mathcal{G}}$ comprises edge-wise operators derived from the geometric product $P_{uv} = \mathbf{H}_u \mathbf{H}_v$. 
    As a bilinear map, the geometric product satisfies the sub-multiplicative perturbation bound:
    \begin{equation}
    \begin{aligned} 
        \|\mathbf{P}_{uv} - \mathbf{P}'_{uv}\|_{\mathcal{C}l} &= \|\mathbf{H}_u \mathbf{H}_v - \mathbf{H}'_u \mathbf{H}'_v\|_{\mathcal{C}l} \\ 
        &\le \|\mathbf{H}_u\|_{\mathcal{C}l} \|\mathbf{H}_v - \mathbf{H}'_v\|_{\mathcal{C}l} + \|\mathbf{H}_u - \mathbf{H}'_u\|_{\mathcal{C}l} \|\mathbf{H}'_v\|_{\mathcal{C}l} \\ 
        &\le \sqrt{K} \cdot \delta_H + \delta_H \cdot \sqrt{K} = 2\sqrt{K} \delta_H.
    \label{eq:geometric_product_stability}
    \end{aligned}
    \end{equation}
    Consequently, the bi-vector term $\mathcal{B}_{uv} = \langle \mathbf{P}_{uv} \rangle_2$ and the scalar term $\mathcal{S}_{uv} = \langle \mathbf{P}_{uv} \rangle_0$ are both bounded by $2\sqrt{K}\delta_H$. 
    The spatial rotor $\mathcal{R}_{uv}$ is derived by normalizing the bi-vector. Assuming the bi-vector norm is bounded away from zero ($\|\mathcal{B}_{uv}\|_{\mathcal{C}l} \ge \epsilon_0$), the rotor satisfies $\|\mathcal{R} - \mathcal{R}'\| \le L_{\mathcal{R}} \delta_H$ due to the Lipschitz continuity of the exponential map on the compact Spin group. 
    Furthermore, the potential $\Phi$ is Lipschitz continuous ($L_{\Phi}$), resulting in $\|\Phi - \Phi'\| \le L_{\Phi} \delta_H$. 
    The Clifford-valued adjacency $\mathcal{A}_{\mathcal{G}}$ is a composite operator $\mathbf{A} \circ (\Phi, \mathcal{R})$. 
    Applying the Leibniz product rule:
\begin{equation}
\begin{aligned}
    \|\mathcal{A}_{\mathcal{G}} - \mathcal{A}'_{\mathcal{G}}\|_{\mathcal{C}l} \le \|\mathbf{A}\|_{\mathcal{C}l} (L_{\mathcal{R}} + L_{\Phi}) \delta_H + C_{\text{struct}} \|\mathbf{A} - \mathbf{A}'\|_{\mathcal{C}l}.
\label{eq:appendix_operator_stability}
\end{aligned}
\end{equation}
\end{lemma}
    Finally, synthesizing Lemma~\ref{theorem_proof_lemma1} and Lemma~\ref{theorem_proof_lemma2} into the total manifold distance $\mathcal{D}_{\mathcal{M}}$:
\begin{equation}
\begin{aligned}
    \mathcal{D}_{\mathcal{M}} &\le \delta_H + \left( \|\mathbf{A}\|_{\mathcal{C}l} (L_{\mathcal{R}} + L_{\Phi}) \delta_H + C_{\text{struct}} \|\mathbf{A} - \mathbf{A}'\|_{\mathcal{C}l} \right) \\
    &= (1 + \|\mathbf{A}\|_{\mathcal{C}l} (L_{\mathcal{R}} + L_{\Phi})) L_{\psi} \epsilon_X + C_{\text{struct}} \epsilon_A.
\end{aligned}
\end{equation}
    By defining the constants $K_1 = (1 + \|\mathbf{A}\|_{\mathcal{C}l} (L_{\mathcal{R}} + L_{\Phi})) L_{\psi}$ and $K_2 = C_{\text{struct}}$, we arrive at the final stability bound:
\begin{equation}
\label{theorem1_proof}
\begin{aligned}
    &\;\;\;\;\| \mathbf{H} - \mathbf{H}' \|_{\mathcal{C}l} + \| \mathcal{A}_{\mathcal{G}} - \mathcal{A}'_{\mathcal{G}} \|_{\mathcal{C}l} \le\\ 
    &K_{{map}} \cdot \left( \|\mathbf{X} - \mathbf{X}'\|_{\mathcal{C}l} + \gamma \|\mathbf{A} - \mathbf{A}'\|_{\mathcal{C}l} \right),
\end{aligned}
\end{equation}
    where $K_{\text{map}} = \max(K_1, K_2)$ and $\gamma$ is the structural weighting factor.
    This inequality rigorously demonstrates that the construction of the Clifford geometric manifold satisfies Lipschitz continuity. 
    Consequently, the geometric descriptors generated by $f$ are theoretically guaranteed to be robust against input perturbations, providing a stable foundation for the subsequent parallel-transport-based propagation.
\end{proof}
    In conclusion, the proof of Theorem~\ref{theorem1} not only confirms that the mapping $f$ from Euclidean space to the curved Clifford manifold is Lipschitz continuous but also rigorously establishes its well-posedness by deriving an explicit stability bound $K_{map}$.
    This mathematical stability ensures that the LION paradigm possesses bounded sensitivity to input noise, where minor fluctuations in modality attributes $\mathbf{X}$ or structural anomalies in the graph topology $\mathbf{A}$ are strictly confined within a constant factor determined by the intrinsic algebraic properties of Clifford operators. 
    Such robustness is indispensable for the manifold's construction, as it safeguards the structural integrity of the modality-aware orthogonal basis vectors and the numerical stability of curvature operators against error accumulation during high-order propagation.

\section{Proof of Spectral Evidence in Theorem~\ref{theorem2}}
\label{appendix: Proof of Spectral Evidence}
    Building upon the stable geometric manifold established in Theorem~\ref{theorem1}, we elucidate the mechanism through which LION achieves modality alignment. 
    Unlike conventional GNNs that rely on scalar homophily smoothing, which frequently induces feature collapse, LION facilitates alignment via CGP.
    By analyzing this process through the lens of graph signal processing, we demonstrate that CGP is not a heuristic smoothing operation but a rigorous optimization algorithm over the curved manifold. 
    Specifically, the layer-wise update defined in Eq.~\ref{eq: Training-free Geometric Propagation} inherently minimizes the potential-gated Clifford Dirichlet energy, which quantifies the aggregate alignment cost across the graph.
    In this section, we provide the rigorous proof for Theorem~\ref{theorem2}, demonstrating how this minimization reduces the geometric divergence between connected neighbors following parallel transport. 
    Furthermore, we elucidate how this geometric optimization fundamentally differs from traditional scalar smoothing by preserving the orthogonal structure of multimodal interactions, thereby ensuring efficient modality alignment strictly governed by the manifold geometry.

    \begin{proof}
    Unlike standard graph signal processing, which assumes that connected nodes should exhibit identical scalar features, our framework clarifies that connected nodes achieve alignment only when their respective representations are parallel following a transport between their local tangent spaces.
    Accordingly, we define the Clifford Dirichlet energy $\mathbb{E}_{\text{Dir}}(\mathbf{H})$ as the weighted sum of geometric transport residuals:
    \begin{equation}
    \mathbb{E}_{\text{Dir}}(\mathbf{H}) = \frac{1}{2} \sum_{u \in \mathcal{V}} \sum_{v \in \mathcal{N}(u)} \Phi_{uv} \left\| \mathbf{H}_u - \mathcal{T}_{uv}(\mathbf{H}_v) \right\|_{\mathcal{C}l}^2,
    \end{equation}
    where $\mathcal{T}_{uv}(\mathbf{H}_v) = \mathcal{R}_{uv} \mathbf{H}_v \mathcal{R}_{uv}^{-1}$ and $\Phi_{uv}$ collectively constitute the potential-gated parallel transport operation. 
    Specifically, this operation is jointly driven by the spatial rotor $\mathcal{R}_{uv}$ (which aligns the geometric orientation via curvature) and the geometric potential $\Phi_{uv}$ (which encapsulates both intra- and inter-modality interaction ratios).
    This energy functional $\mathbb{E}_{\text{Dir}}(\mathbf{H})$ quantifies the total geometric misalignment across the manifold.
    
    Building upon this definition, we demonstrate that the CGP layer update constitutes a gradient descent step that minimizes $\mathbb{E}_{\text{Dir}}(\mathbf{H})$. 
    By considering the gradient of the energy function with respect to the multi-vector representation $\mathbf{H}$ and leveraging the properties of the Clifford inner product $\langle x_u x_v \rangle_{\mathcal{C}l}$, the derivative is formulated as follows:
\begin{equation}
    \begin{aligned}
        \nabla_{\mathbf{H}_u} \mathbb{E}_{\text{Dir}}(\mathbf{H}) &= \sum_{v \in \mathcal{N}(u)} \Phi_{uv} \frac{\partial}{\partial \mathbf{H}_u} \left\| \mathbf{H}_u - \mathcal{R}_{uv} \mathbf{H}_v \mathcal{R}_{uv}^{-1} \right\|_{\mathcal{C}l}^2 \\
        &= 2 \sum_{v \in \mathcal{N}(u)} \Phi_{uv} \left( \mathbf{H}_u - \mathcal{R}_{uv} \mathbf{H}_v \mathcal{R}_{uv}^{-1} \right).
    \end{aligned}
\end{equation}
    To find the optimal alignment state $\mathbf{H}^*$, we set the gradient to zero ($\nabla_{\mathbf{H}_u} \mathbb{E}_{\text{Dir}} = 0$):
    \begin{equation}
    \begin{aligned}
            \left( \sum_{v \in \mathcal{N}(u)} \Phi_{uv} \right) \mathbf{H}^*_u = \sum_{v \in \mathcal{N}(u)} \Phi_{uv} \left( \mathcal{R}_{uv} \mathbf{H}^*_v \mathcal{R}_{uv}^{-1} \right).
    \end{aligned}
\end{equation}
    Let $\hat{\Phi}_{uv} = \Phi_{uv} / \sum_{k \in \mathcal{N}(u)} \Phi_{uk}$ denote the normalized geometric potential. Rearranging the aforementioned expression yields the following fixed-point iteration:
    \begin{equation}
    \begin{aligned}
            \mathbf{H}^*_u = \sum_{v \in \mathcal{N}(u)} \hat{\Phi}_{uv} \left( \mathcal{R}_{uv} \mathbf{H}^*_v \mathcal{R}_{uv}^{-1} \right).
    \end{aligned}
\end{equation}
    This derivation rigorously confirms that the CGP propagation rule introduced in Eq.~\ref{eq: Training-free Geometric Propagation} explicitly performs an iterative optimization to minimize $\mathbb{E}_{\text{Dir}}(\mathbf{H})$, thereby guiding the graph toward a state of geometric alignment.
    
    Subsequently, to quantify the convergence rate, we introduce the potential-induced Clifford geometric Laplacian $\mathcal{L}_{\mathcal{G}}$, which is defined by its operator action on the graph signal $\mathbf{H}$:
    \begin{equation}
    \begin{aligned}
            \mathcal{L}_{\mathcal{G}} \mathbf{H}_u = \sum_{v \in \mathcal{N}(u)} \hat{\Phi}_{uv} \left( \mathbf{H}_u - \mathcal{T}_{uv}(\mathbf{H}_v) \right).
    \end{aligned}
\end{equation}
    The propagation process can be reformulated in matrix operator form as $\mathbf{H}^{(l)} = (\mathbf{I} - \mathcal{L}_{\mathcal{G}}) \mathbf{H}^{(l-1)}$.
    Let $\tilde{\mathcal{L}}_{\mathcal{G}} = \mathbf{D}_{\Phi}^{-1/2} \mathcal{L}_{\mathcal{G}} \mathbf{D}_{\Phi}^{1/2}$ denote the symmetric normalized Laplacian. 
    According to spectral graph theory, the contraction of the misalignment error $\mathbf{E}^{(l)} = \mathbf{H}^{(l)} - \mathbf{H}^*$ is governed by the smallest non-zero eigenvalue, or spectral gap, $\lambda_{\min}$ of $\tilde{\mathcal{L}}_{\mathcal{G}}$. Consequently, the convergence rate is strictly bounded as follows:
    \begin{equation}
    \begin{aligned}
            \left\| \mathbf{H}^{(l)} - \mathbf{H}^* \right\|_{\mathcal{C}l} \le (1 - \lambda_{\min}(\tilde{\mathcal{L}}_{\mathcal{G}}))^l \left\| \mathbf{H}^{(0)} - \mathbf{H}^* \right\|_{\mathcal{C}l}.
    \end{aligned}
\end{equation}
    This theoretically guarantees efficient convergence. 
    The term $(1 - \lambda_{\min}(\tilde{\mathcal{L}}_{\mathcal{G}}))$ implies that alignment occurs rapidly in regions characterized by high geometric potential and strong semantic consensus, effectively suppressing noise from unreliable connections while preserving consistent structural patterns.
    
    Finally, we distinguish our approach from traditional homophily-based smoothing. 
    Standard GNNs minimize Euclidean energy, $\sum \|x_u - x_v\|^2$, which forces features to converge toward a common mean vector ($x_u \to x_v$) and inevitably leads to insufficient modality interaction and over-smoothing issues. 
    In contrast, LION minimizes the transport energy within a high-dimensional Clifford manifold where modalities are initialized as orthogonal basis vectors $\{e_k\}_{k=1}^K$. 
    Notably, the parallel transport operation $\mathcal{T}_{uv}(\cdot)$ performs a geometric rotation via the sandwich product. 
    Since the spatial rotors $\mathcal{R}_{uv}$ belong to the Spin group, they are Clifford isometries that preserve the inner product structure of the basis vectors:
    \begin{equation}
    \begin{aligned}
        \left\langle \mathcal{R}_{uv} e_i \mathcal{R}_{uv}^{-1}, \mathcal{R}_{uv} e_j \mathcal{R}_{uv}^{-1} \right\rangle = \langle e_i, e_j \rangle = \delta_{ij}.
    \end{aligned}
\end{equation}
    This implies that while the semantic frames of nodes $u$ and $v$ are aligned to minimize transport divergence, the orthogonality of the interaction channels is strictly preserved.
    The complex semantic curvature, encoded in $\mathcal{R}$, and the connection strength, encoded in $\Phi$, facilitate alignment without collapsing distinct modalities into a single scalar value. 
    Consequently, LION effectively achieves high-order modality alignment while precluding semantic collapse.
    \end{proof}

    This section provides the spectral evidence for Theorem~\ref{theorem2}, establishing that CGP is a rigorous optimization process over the curved manifold rather than a heuristic smoothing operation. 
    By formulating the Clifford Dirichlet energy $\mathbb{E}_{Dir}(H)$ through weighted geometric transport residuals, the proof demonstrates that the iterative update rule in CGP constitutes a gradient descent step that minimizes the aggregate alignment cost across the graph. 
    The convergence rate of this process is strictly bounded by the spectral gap of the potential-induced geometric Laplacian $\tilde{\mathcal{L}}_{\mathcal{G}}$, ensuring that the alignment error contracts exponentially as semantic consensus is reached. 
    Unlike conventional Euclidean-based GNNs that risk feature collapse and over-smoothing by forcing features toward a common mean, LION leverages the isometric properties of spatial rotors within the Spin group to perform precise geometric rotations. 
    This strategy ensures that the semantic frames are aligned to minimize transport divergence while the fundamental orthogonality of multimodal interaction channels is strictly preserved, thereby providing a robust theoretical guarantee for high-order modality alignment without semantic collapse.
    
\section{Proof of Holographic Reconstruction in Theorem~\ref{theorem3}}
\label{appendix: Proof of Holographic Reconstruction}
    In this section, we provide the rigorous theoretical analysis for the AHA.
    As discussed in the second Solution and Evaluation in Sec.~\ref{sec: Introduction}, the core motivation of AHA is to reveal the intricate dependencies among aligned tokens to unleash their representation potential.
    Unlike conventional aggregation methods that assume uniform information density, AHA achieves adaptive modality fusion within the geometric manifold space, effectively functioning as a dynamic holographic reconstruction process.
    It selectively reconstructs the optimal node representation $\mathbf{Z}^*$ by filtering geometric grade noise and reconciling topology scale discrepancies.
    Here, we formally prove Theorem~\ref{theorem3}, demonstrating that the reconstruction error of AHA is strictly bounded by the residual grade noise and the consensus profile divergence.

\begin{proof}
    To establish a rigorous foundation for the subsequent proof, we first define the necessary notations and operator properties. 
    Let $f_{\text{cll}}: \mathcal{C}l^K \to \mathbb{R}^d$ denote the Clifford linear layer function, which serves as the projection mapping from the high-dimensional geometric manifold back to the Euclidean semantic space. 
    Under this mapping, the final aggregated node representation $\mathbf{Z}_u$ is formulated as the projection of the resonance-weighted sum of multi-scale features:
\begin{equation}
    \begin{aligned}
        \mathbf{Z}_u = f_{\text{cll}}\left(\sum_{l=0}^{L} \beta_{u,l} \tilde{\mathbf{H}}_u^{(l)}\right),
    \end{aligned}
\end{equation}
    where $\beta_{u,l}$ represents the scale-aware resonance score at depth $l$. 
    For the error analysis, we decompose the reconstruction error $\|\mathbf{Z}_u - \mathbf{Z}^*\|$, where $\mathbf{Z}^*$ denotes the optimal target representation, into two distinct sources of perturbation: (1) filtering loss from noise suppression, and (2) geometric deviation from structural discrepancy. 
    Notably, to bridge the metric spaces, we assume $f_{\text{cll}}$ satisfies the Lipschitz continuity condition with constant $\omega$. 
    This constant serves as a proxy for the transformation's sensitivity to the underlying manifold curvature, implying that the Euclidean distance after projection is strictly bounded by the geometric distance on the Clifford manifold:
\begin{equation}
    \begin{aligned}
        \|f_{\text{cll}}(\mathbf{X}) - f_{\text{cll}}(\mathbf{Y})\|_2 \le \omega \|\mathbf{X} - \mathbf{Y}\|_{\mathcal{C}l}, \quad \forall \mathbf{X}, \mathbf{Y} \in \mathcal{C}l^K.
    \end{aligned}
\end{equation}
    The objective is to bound the norm $\|\mathbf{Z}_u - \mathbf{Z}^*\|$. 
    We begin by applying the triangle inequality to decompose the total error into the filtering loss and the consensus divergence. 
    Let $\mathbf{Z}_u^{\text{raw}} = f_{\text{cll}}(\sum_{l=0}^{L} \beta_{u,l} \mathbf{H}_u^{(l)})$ denote the hypothetical representation derived from the raw, noisy input without filtering. 
    By introducing this intermediate term, the total reconstruction error is strictly bounded by:
\begin{equation}
\begin{aligned}
        \|\mathbf{Z}_u - \mathbf{Z}^*\| &= \|\mathbf{Z}_u - \mathbf{Z}_u^{\text{raw}} + \mathbf{Z}_u^{\text{raw}} - \mathbf{Z}^*\| \\
        &\le \underbrace{\|\mathbf{Z}_u - \mathbf{Z}_u^{\text{raw}}\|}_{\text{Filtering Loss}} + \underbrace{\|\mathbf{Z}_u^{\text{raw}} - \mathbf{Z}^*\|}_{\text{Geometric Deviation}}.
\end{aligned}
\label{eq:appendix_decomposition}
\end{equation}
    First, we analyze the filtering loss. 
    This term quantifies the energy of the components removed by the energy-aware grade filtering. 
    Using the Lipschitz continuity of $f_{\text{cll}}$ and the linearity of the aggregation, we expand the difference:
\begin{equation}
\begin{aligned}
    \|\mathbf{Z}_u - \mathbf{Z}_u^{\text{raw}}\| &= \left\| f_{\text{cll}}\left(\sum_{l=0}^{L} \beta_{u,l} \tilde{\mathbf{H}}_u^{(l)}\right) - f_{\text{cll}}\left(\sum_{l=0}^{L} \beta_{u,l} \mathbf{H}_u^{(l)}\right) \right\| \\
    &\le \omega \sum_{l=0}^{L} \beta_{u,l} \left\| \tilde{\mathbf{H}}_u^{(l)} - \mathbf{H}_u^{(l)} \right\|_{\mathcal{C}l}.
\end{aligned}
\end{equation}
    Recall that the filtered feature is $\tilde{\mathbf{H}}_u^{(l)} = \mathbf{H}_u^{(l)} \odot \boldsymbol{\alpha}_u^{(l)}$. 
    Under the additive noise model $\mathbf{H}_u^{(l)} = \mathbf{S}_u^{(l)} + \mathbf{N}_u^{(l)}$, the gate suppresses noise while ideally preserving the signal. 
    The difference thus corresponds to the suppressed noise, weighted by the layer importance $\beta_{u,l}$:
\begin{equation}
\begin{aligned}
    \left\| \mathbf{Z}_u - \mathbf{Z}_u^{\text{raw}} \right\| \le \sum_{l=0}^{L} \beta_{u,l} \omega \left\| (\mathbf{1} - \boldsymbol{\alpha}_u^{(l)}) \odot \mathbf{N}_u^{(l)} \right\|_{\mathcal{C}l}.
\end{aligned}
\label{eq:appendix_filtering_final}
\end{equation}
    Subsequently, we analyze the geometric Deviation $\|\mathbf{Z}_u^{\text{raw}} - \mathbf{Z}^*\|$. 
    We define the optimal representation $\mathbf{Z}^*$ as the ideal projection of the geometric consensus profile $\mathbf{H}_u^{\text{ctx}}$, i.e., $\mathbf{Z}^* = f_{\text{cll}}(\mathbf{H}_u^{\text{ctx}})$. 
    Utilizing the Lipschitz continuity property, the deviation of the raw aggregated signal from the consensus anchor is bounded by:
\begin{equation}
\begin{aligned}
        \left\| \mathbf{Z}_u^{\text{raw}} - \mathbf{Z}^* \right\| &= \left\| f_{\text{cll}}\left(\sum_{l=0}^{L} \beta_{u,l} \mathbf{H}_u^{(l)}\right) - f_{\text{cll}}(\mathbf{H}_u^{\text{ctx}}) \right\| \\
        &\le \omega \left\| \sum_{l=0}^{L} \beta_{u,l} \mathbf{H}_u^{(l)} - \mathbf{H}_u^{\text{ctx}} \right\|_{\mathcal{C}l}.
\end{aligned}
\end{equation}
    Since the attention weights sum to one ($\sum_{l=0}^{L} \beta_{u,l} = 1$), we can express the consensus profile as a convex combination $\mathbf{H}_u^{\text{ctx}} = \sum_{l=0}^{L} \beta_{u,l} \mathbf{H}_u^{\text{ctx}}$. 
    Substituting this, we rewrite the deviation as the weighted sum of layer-wise differences:
\begin{equation}
\begin{aligned}
    \left\| \mathbf{Z}_u^{\text{raw}} - \mathbf{Z}^* \right\| &\le \omega \left\| \sum_{l=0}^{L} \beta_{u,l} (\mathbf{H}_u^{(l)} - \mathbf{H}_u^{\text{ctx}}) \right\|_{\mathcal{C}l} \\
    &\le \sum_{l=0}^{L} \beta_{u,l} \omega \left\| \mathbf{H}_u^{(l)} - \mathbf{H}_u^{\text{ctx}} \right\|_{\mathcal{C}l}.
\end{aligned}
\label{eq:appendix_geometric_final}
\end{equation}
    Finally, by substituting Eq.~(\ref{eq:appendix_filtering_final}) and Eq.~(\ref{eq:appendix_geometric_final}) back into the decomposition in Eq.~(\ref{eq:appendix_decomposition}), we arrive at the unified inequality stated in Theorem~\ref{theorem3}:
\begin{equation}
\begin{aligned}
    \left\| \mathbf{Z}_u - \mathbf{Z}^* \right\| \le & \sum_{l=0}^{L} \beta_{u,l} \underbrace{\omega \left\| (\mathbf{1} - \boldsymbol{\alpha}_u^{(l)}) \odot \mathbf{N}_u^{(l)} \right\|_{\mathcal{C}l}}_{\text{Noise Suppression}} \\
    & + \sum_{l=0}^{L} \beta_{u,l} \underbrace{\omega \left\| \mathbf{H}_u^{(l)} - \mathbf{H}_u^{\text{ctx}} \right\|_{\mathcal{C}l}}_{\text{Scale Consensus}}.
\end{aligned}
\end{equation}
\end{proof}
    In conclusion, this proof theoretically validates the design rationale of the two core components within AHA, establishing a rigorous link between our algebraic construction and the model's robustness. 
    The first term confirms that energy-aware grade filtering minimizes the noise upper bound by suppressing irrelevant or distortion-prone geometric grades. 
    The second term demonstrates that scale-aware resonance fusion minimizes structural deviation by ensuring the final representation converges toward the consensus profile anchor $\mathbf{H}_u^{\text{ctx}}$.

\section{Theoretical Foundations of LION}
\label{appendix: Theoretical Foundations of LION}
    In the above sections, we have established a comprehensive theoretical framework that rigorously validates the mathematical soundness of LION: Theorem~\ref{theorem1} ensures the geometric stability of the Clifford manifold construction against input perturbations; Theorem~\ref{theorem2} proves that high-order propagation achieves modality alignment through potential-gated Dirichlet energy minimization; and Theorem~\ref{theorem3} validates that adaptive aggregation minimizes the holographic reconstruction error for optimal modality fusion. 
    This analysis confirms that LION is not a heuristic architecture but a rigorous geometric optimization process, offering provable guarantees for convergence and representational fidelity.
    
    Specifically, Appendix~\ref{appendix: Proof of Geometric Stability} establishes the geometric stability of the Clifford manifold construction. 
    By proving that the mapping function satisfies the Lipschitz continuity condition, we demonstrate that the projection from Euclidean space to the Clifford manifold is well-posed and inherently robust to input perturbations. 
    This theoretical guarantee ensures that the high-dimensional geometric descriptors serve as a stable foundation.
    Based on this, the spectral evidence detailed in Appendix~\ref{appendix: Proof of Spectral Evidence} reveals that CGP functions as a rigorous gradient descent optimization rather than simple smoothing.
    We theoretically prove that the layer-wise update iteratively minimizes the potential-gated Clifford Dirichlet energy, guiding the graph toward geometric alignment. 
    Notably, unlike traditional scalar GNNs that rely on homophily and risk feature collapse, this mechanism strictly preserves the orthogonality of interaction channels. 
    This ensures that modality alignment is achieved through precise geometric rotation guided by manifold curvature, thereby preventing the over-smoothing issues prevalent in existing methods.
    Finally, the holographic reconstruction bound provided in Appendix~\ref{appendix: Proof of Holographic Reconstruction} validates the design of AHA.
    By strictly bounding the reconstruction error via noise suppression and structural alignment terms, we confirm that our specific design choices are mathematically optimal.
    The proof demonstrates that energy-aware grade filtering effectively minimizes the noise upper bound by suppressing distortion-prone grades, while scale-aware resonance fusion minimizes structural deviation. 
    This guarantees that the final representation converges to the consensus profile with maximal information fidelity.
    These theoretical results collectively substantiate the effectiveness of LION, providing a solid mathematical basis for the algorithmic implementation and complexity analysis detailed in the subsequent section.

\section{Algorithm and Complexity Analysis}
\label{appendix: Algorithm and Complexity Analysis}

\textbf{Algorithm Details}.
    For a more comprehensive presentation, we provide the complete LION in Algorithm~\ref{alg: lion}.
    Specifically, our framework transforms raw multimodal inputs into a unified representation space via two decoupled phases: Clifford Geometric Propagation (CGP) for modality alignment and Adaptive Holographic Aggregation (AHA) for modality fusion.
    
    The data flow begins with the initialization of the Clifford geometric manifold. 
    Given a multimodal-attributed graph $\mathcal{G}$, we employ a modality-oriented isomorphic lifting map to project raw scalar features $\mathbf{X}$ into the Grade-1 subspace of the Clifford algebra $\mathcal{C}l_{K}$. 
    By assigning each modality to a unique orthogonal basis vector $e_k$, this step preserves the independence of interaction channels while unifying them within a multi-vector space to yield the initial representation $\mathbf{H}^{(0)}$. 
    Meanwhile, we retain the remaining grade components for subsequent intra- and inter-modality interactions. 
    After that, to model the topology, we compute the geometric product between connected nodes, decomposing interactions into a scalar component representing projection intensity and a bi-vector component representing the interaction plane. 
    These elements instantiate the topology-oriented geometric potential $\Phi$ and the spatial rotor $\mathcal{R}$, which collectively encode the manifold's semantic curvature. 
    Guided by these operators, we execute training-free high-order propagation. 
    In each layer, node representations undergo potential-gated parallel transport: the spatial rotor aligns the geometric orientation of neighbor tokens with the target node's local tangent space, while the geometric potential provides alignment principles based on semantic consensus. 
    This process iteratively minimizes the Clifford Dirichlet energy, generating a sequence of aligned, curvature-adaptive tokens $\{\mathbf{H}^{(1)}, \dots, \mathbf{H}^{(L)}\}$ that capture multiscale context while preventing semantic collapse.
    
    To provide a more intuitive understanding, we illustrate the process using text and vision modalities as examples below (i.e., $K=2$).
    In this setting, the modality-attributed vector for any node $u$ in the MAG is denoted by $x_u \in \mathbb{R}^{d_t + d_i}$ where $d_t$ and $d_i$ represent the initial embedding dimensions for text and image modalities, respectively, with $d = d_t + d_i$. 
    
    In CGP initialization, we employ our proposed modality-oriented isomorphic lifting map to transform $x_u$ into $\mathbf{H}_u^{(0)} \in \mathbb{R}^{d \times 4}$, which comprises three geometric grades. 
    Specifically, Grade-0 is initialized as a zero vector in $\mathbb{R}^{d}$ to subsequently capture intra-modality interactions. 
    Grade-1 corresponds to the original scalar features in $\mathbb{R}^{d \times 2}$ and serves as the basis for modality interactions. 
    Grade-2 is likewise initialized as a zero vector in $\mathbb{R}^{d}$ to facilitate subsequent inter-modality interactions.
    
    After $l$-step CGP, the multi-vector remains $\mathbf{H}_u^{(l)} \in \mathbb{R}^{d \times 4}$ though the physical interpretation of its geometric components evolves. 
    While Grade-1 technically exists, its feature channels no longer represent original scalar features but are instead subsumed by Grade-0 and Grade-2. 
    Specifically, Grade-0 inherits half of the dimensions from Grade-1 to represent text-text and image-image interaction channels with a total dimension of $\mathbb{R}^{d \times 2}$. 
    Similarly, Grade-2 inherits the remaining half of the Grade-1 dimensions to represent text-image and image-text interaction channels in $\mathbb{R}^{d \times 2}$. 
    This evolution of the physical significance of propagated features within CGP directly motivates our subsequent AHA module.

    Following the propagated multi-vector features naturally exhibit grade-wise properties, where distinct geometric grades encode specific intra- and inter-modality interactions.
    (1) Energy-aware Grade Filtering:
    AHA first quantifies the information density of each token by computing the Clifford norm energy across Grade-0 and Grade-2. 
    Notably, the geometric grades considered here correspond exclusively to Grade-0 and Grade-2 as they serve as the core components for capturing intricate multimodal interactions.
    As previously established, these components precisely correspond to the exhaustive set of $2^2$ interaction channels encompassing all modalities (i.e., $\mathbf{H}_u^{(l)} \in \mathbb{R}^{d \times 4}$).
    While this quantification highlights channels with high signal intensity, high energy does not invariably equate to task relevance. 
    Consequently, we introduce a learnable gating mechanism to govern the adaptive fusion. 
    This gate functions as a semantic filter optimized for the downstream objective, enabling the model to autonomously suppress noise or retain low-energy channels that may carry subtle yet critical interaction signals, rather than relying solely on raw energy magnitude.
    (2) Scale-aware Resonance Fusion:
    Subsequently, to reconcile the varying receptive fields inherent in layer-wise aligned tokens, we employ a scale-aware adaptive fusion mechanism.
    In graph and multimodal learning, different propagation depths capture correlations ranging from local nuances to global structures. 
    To identify the optimal scope, the consensus profile $\mathbf{H}^{\text{ctx}}$ is synthesized to serve as a semantic anchor. 
    This approach is grounded in the rationale that the consensus profile, aggregated from diverse perspectives, provides the most robust approximation of the underlying semantic intent. 
    By calculating resonance scores against this anchor, the model adaptively weights each propagation depth, prioritizing the receptive field that exhibits the highest consistency with the consensus profile.
    Finally, the node representation is obtained by aggregating these filtered, scale-weighted tokens and projecting the resulting multi-vector back into Euclidean space via a Clifford linear transformation. 
    This holographic reconstruction ensures that the output $\mathbf{Z}$ optimally integrates the complex topology and modality dependencies captured within the geometric manifold.

\begin{algorithm}[t]
\caption{LION: c\underline{LI}ff\underline{O}rd \underline{N}eural Paradigm for Multimodal-Attributed Graphs}
\label{alg: lion}
\begin{algorithmic}[1]
\REQUIRE Multimodal-Attributed Graph $\mathcal{G}=(\mathcal{V}, \mathcal{E}, \{\mathbf{X}^{(m)}\}_{m\in\mathcal{M}})$, Model Layers $L$ (Graph Propagation Depth).
\ENSURE Learned Node-level or Modality-level Representations $\mathbf{Z}$.

\STATE \textcolor{blue}{{/* Step 1: Clifford Geometric Propagation for Modality Alignment}}
\STATE \textcolor{blue}{// CGP obtains graph propagated features, which are equivalent to generating aligned tokens by modality alignment.}
\STATE Lift Euclidean modality-attributed features $\mathbf{X}$ into the Clifford manifold to obtain $\mathbf{H}^{(0)}$ via Eq.~(\ref{eq: modality-oriented Clifford initialization}).
\STATE Capture intricate semantic curvature to instantiate $\mathcal{A_G}\!\coloneqq\!\{\Phi,\mathcal{R}\}$ and then facilitate multimodal interactions via Eq.~(\ref{eq: topology-oriented Clifford initialization}). 
\STATE Perform modality-aware, curvature-adaptive high-order graph propagation to obtain $\{\mathbf{H}^{(1)},\dots,\mathbf{H}^{(L)}\}$ via Eq.~(\ref{eq: Training-free Geometric Propagation}).
\STATE \textcolor{blue}{{/* Step 2: Adaptive Holographic Aggregation for Modality Fusion}}
\STATE \textcolor{blue}{// AHA follows CGP to obtain node- or modality-level outputs for specific tasks, which is equivalent to modality fusion.}

\STATE Quantify the information density of modality interaction channels within the propagated features via Eq.~(\ref{eq: energy_energy}).
\STATE Dynamically modulate these channels based on the computed information density and a learnable gate via Eq.~(\ref{eq: energy_gating}).
\STATE \textcolor{blue}{// This module facilitates adaptive integration of modality interaction channels by energy-based information density.}
\STATE Capture the consensus profile across propagation depths via Eq.~(\ref{eq: scale_weight}). 
\STATE Dynamically modulate the contribution of each depth based on the consensus profile via Eq.~(\ref{eq: scale_aggregation}). 
\STATE \textcolor{blue}{// This module facilitates the adaptive integration of multi-scale receptive fields by semantic consensus.}
\STATE \textbf{Return} The optimal fusion representation in Euclidean space.
\end{algorithmic}
\end{algorithm}
\vspace{-0.2cm}

\textbf{Theoretical Time-Space Complexity.} 
    In this section, we rigorously analyze the computational complexity of the LION framework to substantiate its scalability and efficiency for large-scale MAGs.
    To maintain consistency with established benchmarking standards, let $N = |\mathcal{V}|$ and $M = |\mathcal{E}|$ denote the total number of nodes and edges in the MAG, respectively. 
    We define $L$ as the propagation depth and $d$ as the projected latent dimension. Within the Clifford manifold $\mathcal{C}l_K$ involving $K$ distinct modalities, the total dimensionality of a multi-vector is represented as $D = 2^K \cdot d$. 
    LION follows a decoupled alignment-then-fusion paradigm (specifically, a propagation-then-aggregation sequence), effectively partitioning the computational burden into a training-free pre-processing phase and an efficient training phase.
    
    Pre-processing Phase: 
    The CGP phase focuses on extracting high-level semantic representations and capturing global topology dependencies before the start of parameter optimization. 
    Manifold initialization via the isomorphic lifting map $\psi(\cdot)$ requires $\mathcal{O}(N \cdot D)$. 
    The instantiation of the geometric potential $\Phi$ and spatial rotor $\mathcal{R}$ necessitates the calculation of the geometric product $\mathbf{H}_u \mathbf{H}_v$ for each edge, which incurs a cost of $\mathcal{O}(M \cdot 2^K \cdot D)$. 
    Subsequently, the curvature-adaptive high-order propagation scales as $\mathcal{O}(L \cdot M \cdot D)$. 
    Since this phase is training-free and parameter-independent, it can be executed once and cached offline to guide future learning processes.
    
    Training Phase:  
    LION exclusively optimizes the AHA module, significantly reducing the per-epoch computational footprint. 
    Energy-aware grade filtering requires quantifying grade-wise information density across $2^K-1$ subspaces, involving Clifford norm computations of $\mathcal{O}(L \cdot N \cdot D)$. 
    The scale-aware resonance fusion employs an attention mechanism to synthesize the consensus profile $\mathbf{H}_u^{\text{ctx}}$, which also scales as $\mathcal{O}(L \cdot N \cdot D)$. 
    Notably, this phase avoids the recursive neighborhood expansion—a common bottleneck in traditional GNNs—drastically reducing the per-epoch training time. 
    Consequently, the total complexity is $\mathcal{O}(L(M \cdot D + N \cdot 2^K \cdot D))$, which remains asymptotically linear with respect to the graph size $(N, M)$.
    
    Memory Overhead and Practical Scalability:
    LION requires $\mathcal{O}(L \cdot N \cdot D)$ to store the pre-computed multiscale propagated features $\{\mathbf{H}^{(l)}\}_{l=0}^L$ for subsequent aggregation, a design choice that enables the AHA module to reconcile various receptive fields without redundant re-computation. 
    The learnable parameters, primarily residing in the energy gate weight $\mathbf{W}_{\mathcal{G}}$, the consensus profile anchors $\mathbf{W}_{\tau}$, the resonance attention vector $\mathbf{W}_S$, and the output projection layer $\mathbf{W}_{\text{out}}$, occupy $\mathcal{O}(D^2)$ space.
    Although the manifold dimension $D = 2^K \cdot d$ scales exponentially with the number of modalities $K$, this overhead remains marginal in practice as $K$ is typically limited (e.g., $K \in \{2, 3\}$) in most practical multimodal scenarios. 
    
    By executing the CGP as a one-time pre-processing step, the propagated features can be efficiently retrieved during the training of the AHA module. 
    This decoupling enables LION to leverage standard mini-batch sampling techniques without the necessity of loading the dense adjacency matrix into GPU memory.
    Furthermore, by utilizing these pre-computed neighborhood contexts as static inputs, LION entirely precludes the need for recursive neighborhood expansion during back propagation, thereby avoiding the exponential neighbor explosion and the high computational overhead commonly associated with deep message-passing architectures. 
    This decoupled architecture ensures that the training complexity remains asymptotically linear relative to the number of nodes and edges, facilitating high scalability for real-world applications. 
\section{Dataset Description}
\label{appendix: Dataset Description}

\begin{table}[htbp]
\centering
\caption{The statistical information of the experimental datasets.}
\label{table: datasets}
\begin{tabular}{l|cccc|ll}
\midrule[0.3pt]
Datasets     & \# Modalities & \# Nodes & \# Edges   & \# Classes & Tasks                        & Description    \\ \midrule[0.3pt]
RedditS      & Text, Image  & 15,894  & 566,160   & 20              & Graph (Node, Link)           & Social Network \\
Movies       & Text, Image  & 16,672  & 218,390   & 20              & Graph (Node, Link)           & Movie Network  \\
Grocery      & Text, Image  & 17,074  & 171,340   & 20              & Graph (Node, Link), Modality & Recommendation \\
SemArt       & Text, Image  & 21,382  & 1,216,432 & -               & Graph (Link), Modality       & Art Network    \\
Flickr30k    & Text, Image  & 31,783  & 181,151   & -               & Graph (Link), Modality       & Image Network  \\
Sports       & Text, Image  & 50,250  & 356,202   & -               & Graph (Link), Modality       & Recommendation \\
Ele-fashion  & Text, Image  & 97,766  & 199,602   & 12              & Graph (Node, Link), Modality & Recommendation \\
Cloth        & Text, Image  & 125,839 & 951,271   & -               & Graph (Link), Modality       & Recommendation \\
Goodreads & Text, Image  & 685,294 & 7,235,048 & 11              & Graph (Node, Link)           & Book Network   \\ \midrule[0.3pt]
\end{tabular}
\vspace{0.3cm}
\end{table}

\textbf{RedditS}~\cite{desai1_RedditS}
    is a social network from Reddit where nodes represent posts and edges denote threading relationships (e.g., comments and replies). 
    Textual features are encoded from titles and body content, while visual features are extracted from embedded images.
    This dataset is primarily used for node classification and node clustering.

\textbf{Movies}~\cite{ni2019_Grocery_Cloth_Ele_Movies_Sports} 
    is sourced from Amazon’s Movies and TV category. 
    Nodes correspond to DVD/Blu-ray products, and edges reflect consumer co-purchasing behavior. 
    Node attributes include textual plot synopses and customer reviews, alongside visual features derived from official cover art.

\textbf{Grocery}~\cite{ni2019_Grocery_Cloth_Ele_Movies_Sports} 
    originates from Amazon’s Grocery and Gourmet Food segment. 
    Edges indicate complementary purchasing habits derived from "also-bought" metadata. 
    Textual features are encoded from product titles and nutritional descriptions, while visual features are extracted from packaging images. 
    This dataset poses the fine-grained node classification and node clustering task focused on product sub-categories.

\textbf{SemArt}~\cite{garcia2018_SemArt} 
    is a fine-art dataset where nodes represent paintings and edges are established based on shared metadata such as artist, period, or school. 
    Node features include expert historical commentary and stylistic visual attributes from digital images. 
    SemArt serves as a benchmark for Graph-to-Image (G2Image) tasks, requiring the reconciliation of abstract historical descriptions with complex visual aesthetics.

\textbf{Flickr30k}~\cite{plummer2015_Flickr30k} 
    is a canonical image-text reasoning dataset. 
    In the OpenMAG setting, we construct a graph where nodes represent image regions and caption phrases, linked by semantic grounding annotations. 
    This dataset is utilized for Graph-to-Text (G2Text) tasks, evaluating the model's ability to generate descriptive captions by traversing grounded visual-textual relationships.

\textbf{Sports}~\cite{hou2024_Cloth_Ele_Sports, ni2019_Grocery_Cloth_Ele_Movies_Sports} 
    is an Amazon-based graph of athletic gear where edges denote functional complementarity.
    It incorporates technical specifications (text) and product images (visual) to capture design and utility. 
    This dataset is primarily utilized for link prediction, aiming to forecast potential product associations by modeling the geometric compatibility between sports-related modalities.

\textbf{Ele-fashion}~\cite{ni2019_Grocery_Cloth_Ele_Movies_Sports,hou2024_Cloth_Ele_Sports} 
    is a heterogeneous graph merging Amazon’s Electronics and Fashion categories. Nodes are connected via cross-category co-purchasing links, revealing latent consumer preferences across disparate domains.
    Features combine technical specs with style descriptions and product imagery.

\textbf{Cloth}~\cite{ni2019_Grocery_Cloth_Ele_Movies_Sports,hou2024_Cloth_Ele_Sports} 
    is a large-scale Amazon fashion graph linking items via visual compatibility or co-purchase history. 
    It utilizes material compositions (text) and high-resolution model images (visual).
    The dataset is used for node classification and node clustering of fashion styles, where high intra-class variance and subtle inter-class differences provide a robust test for modality fusion and alignment.

\textbf{Goodreads}~\cite{wan2018_Goodreads_NC,wan2019_Goodreads_NC} 
    is a book review graph where edges represent co-shelving relationships. 
    Node features are derived from book summaries, editorial reviews, and cover art.
    The tasks are node classification and node clustering of literary genres, requiring the synthesis of aesthetic cover cues with the semantic depth of textual summaries to achieve accurate classification.

\section{Baselines Details}
\label{appendix: Baselines Details}

    \textbf{GCN}~\cite{kipf2016gcn} utilizes a localized first-order approximation of spectral graph convolutions based on the renormalization trick.
    It scales linearly with the number of edges and learns hidden representations that jointly capture graph topology structure and node attribute features via a layer-wise propagation rule.
    
    \textbf{GCNII}~\cite{chen2020gcnii} extends the conventional GCN by incorporating two key techniques: initial residual connections and identity mapping.
    These mechanisms effectively simulate lazy random walks and are theoretically and empirically shown to alleviate the over-smoothing issue, allowing the model to stack deep architectures without performance degradation.
    
    \textbf{GAT}~\cite{velivckovic2017gat} introduces masked self-attention layers to assign learnable importance scores to neighbors during aggregation.
    This strategy enables the model to handle anisotropic graphs by implicitly assigning varying weights to different nodes, thereby capturing complex local structural patterns without relying on a fixed, predefined graph Laplacian.
    
    \textbf{GATv2}~\cite{brody2021gatv2} implements a dynamic graph attention mechanism that strictly improves upon the static attention of the original GAT.
    By modifying the order of operations in the attention scoring function, it achieves universal approximation capability, offering enhanced representation expressiveness and increased robustness against graph noise and irrelevant connections.
    
    \textbf{MMGCN}~\cite{hu2021mmgcn} is a pioneering framework for micro-video recommendation that explicitly models user preferences across visual, acoustic, and textual modalities.
    It constructs separate modality-specific bipartite graphs and aggregates high-order connectivity within each graph before combining them through a structured fusion layer, effectively capturing the user-item interactions inherent in each sensory channel.
    
    \textbf{MGAT}~\cite{tao2020mgat} employs a gated attention mechanism within parallel multimodal interaction graphs for recommendation.
    It adaptively identifies the importance of specific modalities to disentangle granular personal interests, effectively acting as a denoising filter to reduce the influence of noisy or conflicting multimodal signals during preference learning.
    
    \textbf{MLaGA}~\cite{fan2025mlaga} enables Large Language Models (LLMs) to reason over MAGs via a coherent two-stage alignment strategy.
    It first aligns visual and textual features with the graph structure through contrastive graph pre-training, and subsequently performs instruction tuning to seamlessly integrate graph connectivity priors into the LLM's generative reasoning process.
    
    \textbf{GraphGPT-O}~\cite{fang2025graphgpt_o} is a comprehensive multimodal LLM designed for joint comprehension and generation tasks on MAGs.
    It addresses the challenge of encoding non-Euclidean dependencies and scalability by employing personalized PageRank sampling and a hierarchical aligner equipped with both node-level and graph-level Q-Formers to bridge structural tokens with the LLM semantic space.
    
    \textbf{Graph4MM}~\cite{ning2025graph4mm} integrates multi-hop structural information directly into the self-attention mechanism via a hop-diffused attention strategy.
    It utilizes a specialized MM-QFormer for principled cross-modal fusion, demonstrating that treating graph topology as a guided interaction modality significantly outperforms approaches that treat the graph merely as an auxiliary input feature.
    
    \textbf{InstructG2I}~\cite{jin2024instructg2i} is a graph-conditioned diffusion model specifically designed for MAGs.
    It utilizes semantic neighbor sampling to construct contextual prompts and employs a Graph-QFormer to encode these prompts, offering fine-grained controllability over the generative process through a novel graph classifier-free guidance mechanism.
    
    \textbf{DMGC}~\cite{guo2025DMGC} addresses complex hybrid neighborhood patterns by explicitly disentangling the graph into cross-modality homophily-enhanced and modality-specific heterophily-aware components.
    It utilizes a dual-frequency fusion mechanism, which acts as coupled low-pass and high-pass filters to simultaneously capture both intra-class commonalities (smoothness) and inter-class distinctions (boundaries).
    
    \textbf{DGF}~\cite{zheng2025dgf} proposes a cross-contrastive clustering framework that utilizes a dual graph filtering scheme to systematically denoise features extracted from MAG.
    It employs a tri-cross contrastive objective spanning modalities, topology neighborhoods, and semantic communities to learn discriminative clustering representations robust to outliers.
    
    \textbf{MIG-GT}~\cite{hu2025mig_gt} utilizes modality-independent GNNs with adaptive receptive fields to accommodate the unique propagation requirements and noise levels of distinct modalities.
    To complement local aggregation, it integrates a sampling-based global transformer, enabling the model to capture long-range semantic dependencies and global context that standard message passing often fails to preserve.
    
    \textbf{NTSFormer}~\cite{hu2025ntsformer} introduces a self-teaching graph transformer tailored for isolated cold-start node classification.
    It employs a stochastic cold-start attention mask to supervise a student prediction (based solely on self-information) with a teacher prediction (which is neighbor-aware), thereby ensuring robust performance and generalization even when structural connections or modal attributes are partially missing.
    
    \textbf{UniGraph2}~\cite{he2025unigraph2} is a cross-domain graph foundation model that integrates diverse modalities into a unified embedding space.
    It leverages frozen pre-trained encoders and a mixture-of-experts (MoE) module for scalable alignment, followed by a universal GNN for structural aggregation, facilitating the learning of generalizable representations that transfer effectively across various downstream tasks and domains.

\section{Evaluation Protocols}
\label{appendix: Evaluation Protocols}
    To ensure a rigorous and standardized evaluation across diverse MAG learning models, we implement specific experimental configurations for three primary graph-centric downstream tasks: supervised node classification, supervised link prediction, and unsupervised node clustering. 
    For node classification and node clustering, the learning rate is established at $5 \times 10^{-3}$ with a batch size of 512 and a weight decay of $1 \times 10^{-5}$. 
    While 100 training epochs are sufficient for convergence in the node classification task, node clustering requires an extended duration of 500 epochs to stabilize the underlying self-supervised objectives. 
    In contrast, the link prediction task utilizes an adjusted learning rate of $1 \times 10^{-3}$ and a significantly larger batch size of 2048 to facilitate the efficient processing of extensive edge-pair samples. 
    To maintain architectural consistency and ensure a fair comparison, we standardize core parameters by employing CLIP-ViT-Large-Patch14 as the default frozen feature encoder, with feature dimensions unified at 768 across all tasks.
    Finally, to mitigate the impact of initialization randomness and provide a robust, unbiased evaluation of model performance, all experiments are repeated ten times to obtain mean results for the reported metrics.
    
    \textbf{Node Classification} is the fundamental supervised task in graph learning. 
    Given a MAG, the model encodes each node into a low-dimensional embedding. 
    These representations are subsequently processed by a projection head and a Softmax layer to generate class probabilities. 
    Optimization is performed by minimizing the cross-entropy loss between the predicted distributions and ground-truth labels.
    Model performance is quantified using Accuracy (Acc) and the F1-score.
    
    Specifically, Acc measures the proportion of correctly predicted samples over the evaluation set: $\text{Acc} = \frac{1}{N} \sum_{i=1}^{N} \mathbb{I}(\hat{y}_i = y_i)$, where $N$ is the sample size, $y_i$ is the ground-truth, $\hat{y}_i$ is the prediction, and $\mathbb{I}(\cdot)$ is the indicator function.
    F1-score is the harmonic mean of Precision and Recall, providing a robust evaluation under class imbalance common in real-world graphs: $\text{F1} = 2 \left(\cdot \text{Precision} \cdot \text{Recall}\right)/\left(\text{Precision} + \text{Recall}\right)$.

    \textbf{Link Prediction}
    measures the model’s capacity to infer missing or potential edges. 
    The model computes similarity scores between node pairs (e.g., via dot product) and assigns higher scores to true edges than to negative samples. 
    In multimodal settings, this requires aligning structural proximity with cross-modal semantic similarity. 
    Evaluation relies on ranking-based metrics: Mean Reciprocal Rank (MRR) and Hits@K.
    
    Specifically, MRR assesses the model's ability to prioritize the correct target at the top of a candidate list: $\text{MRR} = \frac{1}{|Q|} \sum_{i=1}^{|Q|} \frac{1}{\text{rank}_i}$, where $\text{rank}_i$ denotes the rank of the first correct result for query $i$.
    Hits@K reflects the recall performance at a fixed cutoff $K$, indicating the practical utility of the retrieval results: $\text{Hits@K} = \frac{1}{|Q|} \sum_{i=1}^{|Q|} \mathbb{I}(\text{rank}_i \leq K)$.

    \textbf{Node Clustering} evaluates representation quality in an unsupervised setting, following the protocol in \cite{guo2025DMGC}. 
    The model partitions nodes into semantic groups without label supervision. 
    This involves disentangling the graph into homophilous and heterophilous views, followed by a dual-frequency fusion of filtered signals. 
    The model is optimized via a joint objective comprising reconstruction, contrastive alignment, and clustering losses. 
    Model performance is quantified using Normalized Mutual Information (NMI) and Adjusted Rand Index (ARI).
    
    Specifically, NMI quantifies the mutual dependence between predicted clusters $C$ and ground-truth labels $Y$, independent of label permutations: $\text{NMI}(Y, C) = 2 \cdot I(Y, C)/\left(H(Y) + H(C)\right)$, where $I(\cdot)$ and $H(\cdot)$ denote mutual information and entropy, respectively.
    ARI evaluates clustering similarity by considering all sample pairs and adjusting for chance grouping: 
    \begin{equation}
        \begin{aligned}
            \text{ARI} = \frac{\sum_{ij} \binom{n_{ij}}{2} - [\sum_i \binom{a_i}{2} \sum_j \binom{b_j}{2}] / \binom{n}{2}}{ \frac{1}{2} [\sum_i \binom{a_i}{2} + \sum_j \binom{b_j}{2}] - [\sum_i \binom{a_i}{2} \sum_j \binom{b_j}{2}] / \binom{n}{2} },
        \end{aligned}
    \end{equation} 
    where $n_{ij}$ represents the overlap between ground-truth cluster $i$ and predicted cluster $j$.

    For the modality retrieval task, which evaluates the model’s capability to search for relevant instances across image and text modalities within a shared latent space, we employ contrastive learning objectives integrated with a temperature scaling factor of $\tau = 0.07$. 
    These retrieval models are trained for 500 epochs using a learning rate of $1 \times 10^{-3}$ and a batch size of 256. 
    To effectively mitigate over-fitting and ensure the generalizability of the learned representations, we implement an early stopping mechanism with a patience period ranging from 10 to 25 epochs.
    For modality generation tasks, the configurations are specifically tailored to accommodate high-dimensional generative processes and complex many-to-many relationships within the graph. 
    In the Graph-to-Text (G2Text) task, we set the learning rate to $1 \times 10^{-3}$ with a weight decay of $1 \times 10^{-2}$ and a batch size of 8, training the model for 15 epochs.
    We utilize the Self-Attention with Embeddings (SA-E) strategy to sample four multimodal neighbors and employ Graph Neural Networks (GNNs) for structural position encoding. 
    The decoder backbone is powered by the pre-trained Facebook OPT-125M, and to ensure parameter efficiency during adaptation, we support both Prefix Tuning and Low-Rank Adaptation (LoRA) with a rank of $r=64$.
    In the Graph-to-Image (G2Image) task, we employ a learning rate of $1 \times 10^{-4}$ and a batch size of 16 for a duration of 20 epochs.
    Following the InstructG2I framework, we adopt Semantic Personalized PageRank (PPR)-based neighbor sampling, selecting between 0 and 6 informative neighbors to provide structural context. 
    The image resolution is standardized at 256 to condition the Stable Diffusion v1.5 backbone through a Graph Classifier-Free Guidance mechanism.
    
    To ensure experimental fairness and minimize performance variance, we standardize core architectural parameters across all task configurations. 
    Unless otherwise specified, we employ CLIP-ViT-L/14 as the default feature encoder, with node embedding dimensions unified at 768 across all downstream tasks. 
    All optimization processes are performed using the Adam or AdamW optimizers. 
    To further guarantee the robustness of our results and mitigate bias introduced by initialization randomness, we refrain from using fixed random seeds. 
    Each experiment is repeated 5 times, and we report the mean performance across all metrics to provide a comprehensive and unbiased evaluation of the models.

    \textbf{Modality Retrieval} evaluates the model’s capability to retrieve relevant instances across disparate modalities, specifically focusing on image-to-text and text-to-image retrieval within the LION. 
    Given a query from a source modality, the model projects both the query and the candidate set from the target modality into a unified, high-dimensional latent space for direct comparison. 
    Pairwise similarity scores are subsequently computed to rank candidates in descending order of predicted relevance. 
    This task serves as a rigorous assessment of the robustness of cross-modal alignment and the discriminative power of the learned representations for information retrieval. 
    To provide a comprehensive quantitative evaluation, model performance is measured using standard ranking metrics, including Mean Reciprocal Rank (MRR) and Hits@K.

    \textbf{Graph-to-Text (G2Text)} is a generative task requiring the model to synthesize natural language descriptions conditioned on graph-structured multimodal inputs. 
    Diverging from traditional one-to-one multimodal mappings, G2Text addresses complex many-to-many relationships where target nodes interact with diverse multimodal neighborhoods. 
    Following the MMGL framework~\cite{yoon2023mmgl}, the process involves three stages: 
    (i) Neighbor Encoding, which maps diverse modalities into a compatible embedding space; 
    (ii) Graph Structure Encoding, which captures topology context via GNNs or Laplacian position encodings; 
    (iii) Integration, where structure-aware signals are infused into a pre-trained LLM. 
    This task evaluates the faithful translation of structured graph information into coherent text.
    Model performance is evaluated by using standard NLG metrics, including BLEU-4, ROUGE-L, and CIDEr.
    
    Specifically, BLEU-4 evaluates lexical accuracy and fluency by calculating the geometric mean of modified $n$-gram precisions ($p_n$) up to length 4: 
    \begin{equation}
        \begin{aligned}
            \text{BLEU-4} = \text{BP} \cdot \exp\left(\sum_{n=1}^4 w_n \log p_n\right).
        \end{aligned}
    \end{equation}
    ROUGE-L measures sentence-level recall based on the longest common subsequence, ensuring the output covers the comprehensive information content of the ground truth: 
    \begin{equation}
        \begin{aligned}
            \text{ROUGE-L} = \frac{(1 + \beta^2) R_{lcs} P_{lcs}}{R_{lcs} + \beta^2 P_{lcs}}.
        \end{aligned}
    \end{equation}
    CIDEr measures the consensus between generated captions and human references using TF-IDF weighting, emphasizing the semantic importance and distinctiveness of the generated terms:
    \begin{equation}
    \begin{aligned}
        \text{CIDEr}n(c, r) = \frac{1}{M} \sum{i=1}^{M} \frac{g^n(c) \cdot g^n(r_i)}{|g^n(c)| |g^n(r_i)|}.
    \end{aligned}
    \end{equation}

    \textbf{Graph-to-Image (G2Image)} focuses on synthesizing images conditioned on MAG, requiring visual content to reflect both textual prompts and complex graph-based associations. 
    Following the InstructG2I framework~\cite{jin2024instructg2i}, the workflow consists of: 
    (i) Semantic PPR-based Sampling, which selects informative neighbors; 
    (ii) Graph-QFormer Encoding, which transforms neighbors into graph conditioning tokens; 
    (iii) Conditional Generation, where a latent diffusion model is guided by these tokens via a graph classifier-free guidance mechanism. 
    This task evaluates the model's ability to generate images consistent with the styles or categories defined by the graph structure.
    Model performance is quantitatively by using CLIP-Score and DINOv2-Score for instance-level consistency.
    
    Specifically, CLIP-Score quantifies cross-modal semantic consistency. 
    Building upon the textual encoder ($E_T$) and visual encoder ($E_I$) in pre-trained CLIP, this metric determines whether generated images faithfully preserve the semantic content of corresponding graph descriptions:
    \begin{equation}
    \begin{aligned}
        \text{CLIP-Score}(I, T) = \max\left(100 \cdot \cos(E_I(I), E_T(T)), 0\right).
    \end{aligned}
    \end{equation}
    DINOv2-Score assesses visual fidelity and structural consistency using feature embeddings from a pre-trained DINOv2 encoder, ensuring high perceptual quality and structural resemblance to reference samples:
    \begin{equation}
    \begin{aligned}
        \text{DINOv2-Score}(I_{\text{gen}}, I_{\text{ref}}) = \cos(\text{DINO}(I_{\text{gen}}), \text{DINO}(I_{\text{ref}})).
    \end{aligned}
    \end{equation}

\section{Experiment Environment}
\label{appendix: Experiment Environment}
    Experiments are conducted on a workstation equipped with Intel Xeon Scalable processors and NVIDIA RTX 6000 Ada Generation GPUs with 96GB of VRAM, supported by 256GB of system RAM.
    The computational environment utilizes CUDA 12.9, while software implementations are developed using Python 3.10.18 and PyTorch 2.8.
\end{document}